\documentclass[10pt,twocolumn,letterpaper]{article}
\usepackage{times}
\usepackage{xspace}
\usepackage[dvipsnames]{xcolor}
\usepackage{graphicx}
\usepackage{amsmath,amssymb}
\usepackage{booktabs}
\usepackage{colortbl}
\usepackage[numbers,sort&compress]{natbib}
\usepackage[format=plain,labelformat=simple,labelsep=period,font=small,compatibility=false]{caption}
\usepackage[font=footnotesize,skip=3pt,subrefformat=parens]{subcaption}
\usepackage[hyphens]{url}
\usepackage{capt-of}
\usepackage{listings}
\usepackage{placeins}
\definecolor{linkblue}{rgb}{0.21,0.49,0.74}
\usepackage[pagebackref,breaklinks,colorlinks,allcolors=linkblue]{hyperref}
\usepackage[capitalize]{cleveref}
\makeatletter
\def\@maketitle{\newpage\null\vskip .375in
  \begin{center}{\Large\bf \@title\par}\vspace*{24pt}
  {\large\lineskip .5em\begin{tabular}[t]{c}\@author\end{tabular}\par}\vskip .5em\vspace*{12pt}\end{center}}
\def\abstract{\centerline{\large\bf Abstract}\vspace*{12pt}\noindent\it\ignorespaces}

\def\dbsection{\@startsection{section}{1}{\z@}{-10pt plus -2pt minus -2pt}{7pt}{\large\bf}}
\def\dbsubsection{\@startsection{subsection}{2}{\z@}{-8pt plus -2pt minus -2pt}{6pt}{\elvbf}}
\font\elvbf=ptmb7t scaled 1100
\def\section{\@ifstar{\dbsection*}{\dbsect}}
\def\dbsect#1{\dbsection{\texorpdfstring{\hskip -1em.~}{}#1}}
\def\subsection{\@ifstar{\dbsubsection*}{\dbsubsect}}
\def\dbsubsect#1{\dbsubsection{\texorpdfstring{\hskip -1em.~}{}#1}}
\makeatother
\newcommand{\ChainFive}{127}
\newcommand{\ChainFour}{238}
\newcommand{\ChainThree}{489}
\newcommand{\ClaudeOneFortyHZeroHi}{21}
\newcommand{\ClaudeOneFortyHZeroLo}{4}
\newcommand{\DSClaudeHThreeHi}{54}
\newcommand{\DSClaudeHThreeLo}{47}
\newcommand{\DSClaudeHZeroHi}{18}
\newcommand{\DSClaudeHZeroLo}{12}
\newcommand{\DSOpusHThree}{54}
\newcommand{\DSOpusHThreeHi}{60}
\newcommand{\DSOpusHThreeLo}{48}
\newcommand{\DSSmallHThreeHi}{11}
\newcommand{\DSSmallHZeroHi}{2}
\newcommand{\DeathsHi}{3}
\newcommand{\DeathsLo}{0}
\newcommand{\DecideHThreeHi}{86}
\newcommand{\DecideHThreeLo}{82}
\newcommand{\DecideHZeroHi}{61}
\newcommand{\DecideHZeroLo}{57}

\newcommand{\ExactHZeroHi}{47}
\newcommand{\ExactHZeroLo}{39}
\newcommand{\FrontierDecideHThreeHi}{92}
\newcommand{\FrontierDecideHThreeLo}{80}
\newcommand{\FrontierDecideHZeroHi}{75}
\newcommand{\FrontierPairsHZeroHi}{40}
\newcommand{\GFourOneFortyHThreeHi}{66}
\newcommand{\GFourOneFortyHThreeLo}{60}
\newcommand{\GFourOneFortyHZeroHi}{42}
\newcommand{\GFourOneFortyHZeroLo}{31}
\newcommand{\GainOneKDSHi}{36}
\newcommand{\GainOneKDSLo}{26}
\newcommand{\GainOneKHThreeHi}{79}
\newcommand{\GainOneKHThreeLo}{76}
\newcommand{\GainOneKHZeroHi}{53}
\newcommand{\GainOneKHZeroLo}{50}

\newcommand{\HaikuSubHThree}{82}

\newcommand{\LadOneFortyHOneMaxGain}{6}
\newcommand{\LadOneFortyHThreeHi}{91}
\newcommand{\LadOneFortyHThreeLo}{79}
\newcommand{\LadOneFortyHZeroHi}{64}
\newcommand{\LadOneFortyHZeroLo}{57}
\newcommand{\LadOneFortyTxtImgMaxDiff}{6}
\newcommand{\MechItemsOneFortyHi}{20}
\newcommand{\MechItemsOneFortyLo}{14}
\newcommand{\MechItemsOneKHi}{140}
\newcommand{\MechItemsOneKLo}{54}
\newcommand{\MuseHThree}{80}
\newcommand{\MuseHZero}{55}

\newcommand{\NExclFifty}{6}
\newcommand{\NExclOneK}{73}
\newcommand{\NFamilies}{10}
\newcommand{\NFifty}{38}
\newcommand{\NItems}{854}
\newcommand{\NMech}{7}
\newcommand{\NModels}{30}
\newcommand{\NOneForty}{124}
\newcommand{\NOpenFull}{9}
\newcommand{\NPairs}{427}
\newcommand{\NPairsFifty}{19}
\newcommand{\NPairsOneForty}{62}

\newcommand{\NaiveExact}{14}
\newcommand{\NaiveWrong}{731}

\newcommand{\OpenOneKHZeroHi}{53}
\newcommand{\OpenOneKHZeroLo}{25}

\newcommand{\OpusOneFortyDSHThree}{64}
\newcommand{\PairsHThreeHi}{49}
\newcommand{\PairsHThreeLo}{34}
\newcommand{\PairsHZeroHi}{16}
\newcommand{\PairsHZeroLo}{15}
\newcommand{\PeopleHi}{6}
\newcommand{\PeopleLo}{2}

\newcommand{\PerMechOneKHi}{70}
\newcommand{\PerMechOneKLo}{27}
\newcommand{\PerMechOneKSecondLo}{62}

\newcommand{\QwenThreeEightHThree}{82}
\newcommand{\QwenThreeEightHZero}{47}

\newcommand{\RepDSDiffHThree}{15}

\newcommand{\RepDSHThreeA}{64}
\newcommand{\RepDSHThreeB}{49}
\newcommand{\RepDSHZeroA}{8}
\newcommand{\RepDSHZeroB}{14}
\newcommand{\RepDecHThreeA}{91}
\newcommand{\RepDecHThreeB}{86}
\newcommand{\RepDecHZeroA}{59}
\newcommand{\RepDecHZeroB}{67}
\newcommand{\RepSameHThree}{74}
\newcommand{\RepSameHZero}{81}
\newcommand{\SepOne}{89}
\newcommand{\SepThree}{130}
\newcommand{\SepTwo}{208}

\newcommand{\SubsetGap}{5}
\newcommand{\SurviveExact}{25}
\newcommand{\UnparsedQEightInstHThree}{43}
\newcommand{\UnparsedQThirtyAThreeHThree}{69}
\newcommand{\WeightList}{pendulum 0.300, jump 0.250, dominoes 0.152, ledge 0.092, pit 0.073, and recess and seesaw 0.066 each}
\newcommand{\WeightMaxChangeMech}{ledge}
\newcommand{\WeightMaxChangeNew}{0.092}
\newcommand{\WeightMaxChangeOld}{0.118}
\newcommand{\WeightMin}{0.066}

\newcommand{\NonFlipHZeroHi}{90}
\newcommand{\NonFlipHZeroLo}{85}
\newcommand{\RandDSFiftyMean}{8.0}
\newcommand{\RandDSFiftyPct}{25.9}
\newcommand{\RandDSOneFortyMean}{5.6}
\newcommand{\RandDSOneFortyPct}{14.4}
\newcommand{\RandDSOneKMean}{2.6}
\newcommand{\RandDSOneKPct}{6.2}
\newcommand{\SameAnsHThreeHi}{24}
\newcommand{\SameAnsHThreeLo}{22}
\newcommand{\SameAnsHZeroHi}{69}
\newcommand{\SameAnsHZeroLo}{50}
\newcommand{\UniformNFifty}{30}
\newcommand{\UniformNOneForty}{30}
\newcommand{\UniformNOneK}{16}
\newcommand{\UniformRhoFifty}{0.97}
\newcommand{\UniformRhoOneForty}{0.96}
\newcommand{\UniformRhoOneK}{0.95}

\newcommand{\VisDropOneK}{73}

\newcommand{\VisRhoOneKHThree}{0.99}
\newcommand{\VisMeanChangeOneKHThree}{+0.2}
\newcommand{\VisNModelsOneK}{13}
\newcommand{\VisDropOneForty}{8}
\newcommand{\VisRhoOneFortyHZero}{0.90}

\newcommand{\VisRhoOneFortyHThree}{0.96}
\newcommand{\VisMeanChangeOneFortyHThree}{+1.3}
\newcommand{\VisNModelsOneForty}{27}
\newcommand{\VisDropFifty}{6}

\newcommand{\VisRhoFiftyHThree}{0.92}
\newcommand{\VisMeanChangeFiftyHThree}{+10.6}

\newcommand{\VisLedgeOneK}{45}

\newcommand{\CalibGPTConfAccHZeroLo}{59}
\newcommand{\CalibGPTConfAccHZeroHi}{76}
\newcommand{\CalibOpenConfAccHZero}{51}

\newcommand{\CalibConfShareHZeroLo}{63}
\newcommand{\CalibConfShareHZeroHi}{76}
\newcommand{\CalibConfAccHZeroLo}{59}
\newcommand{\CalibConfAccHZeroHi}{63}

\newcommand{\CalibConfAccHThreeLo}{83}
\newcommand{\CalibConfAccHThreeHi}{90}

\newcommand{\SameClosedHZeroLo}{37}
\newcommand{\SameClosedHZeroHi}{76}
\newcommand{\SameClosedHThreeLo}{15}
\newcommand{\SameClosedHThreeHi}{37}
\newcommand{\SameUpModels}{Qwen3-VL-30B-A3B, Qwen2.5-VL-7B and Kimi-VL-A3B-Thinking}

\newcommand{\dbench}{DominoBench}
\title{\dbench: Evaluating Multi-Step Physical Reasoning from Still Images}
\author{Emilio Ferrara\\
University of Southern California\\
{\tt\small emiliofe@usc.edu}}
\begin{document}
\twocolumn[{%
  \renewcommand\twocolumn[1][]{#1}%
  \maketitle
  \begin{center}
  \includegraphics[width=\textwidth]{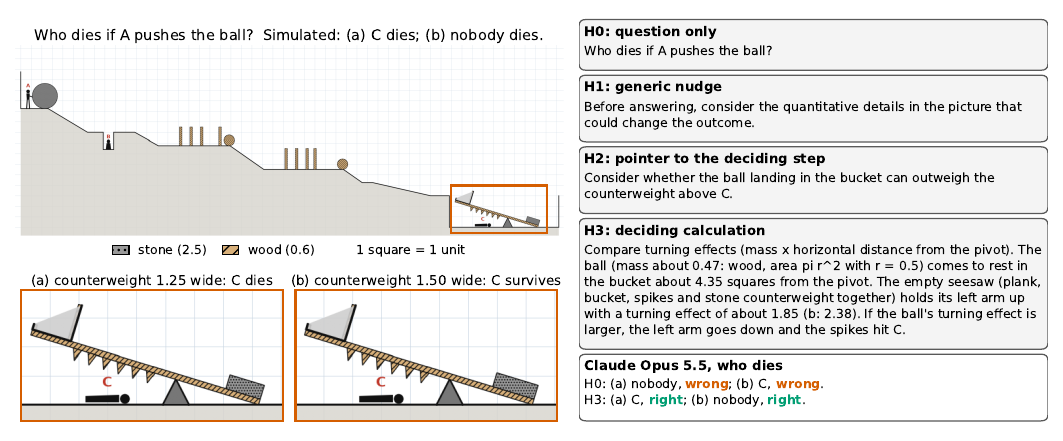}
  \captionof{figure}{A minimal pair from \dbench-1K, also in \dbench-124. Left: the still of item (a). A pushed stone ball sets off two rows of dominoes, and a small ball drops into the bucket of a spiked seesaw above C. Item (b) differs only in a counterweight one grid step (0.25 squares) wider. Zooms: in (a) the ball outweighs the counterweight and the spikes hit C; in (b) the seesaw does not tip. Right: the question and hints, with the H3 value for (b) in parentheses. Each rung is a fresh query with all hints up to it. Unaided, Claude Opus 5.5 gets both items wrong, swapping C's fates; with H3, it gets both right. Data: pair 4 (items 0004-a and 0004-b).}
  \label{fig:teaser}
\end{center}

}]

\begin{abstract}
Can a vision-language model look at a picture of a chain reaction and tell what will happen?
Recent multimodal models describe scenes fluently and solve textbook physics problems, which suggests they should also predict how a scene of ramps, ledges, pendulums, seesaws and dominoes plays out once a ball is pushed.
We tested this claim on \NModels{} models from \NFamilies{} families and reached a clear conclusion: unaided, no model predicts the outcome reliably.
We built \dbench, a benchmark of still images of 2D rigid-body scenes whose outcome a physics simulator computes.
It holds \NItems{} items in \NPairs{} minimal pairs, in which a change of one to three grid steps in one quantity decides whether one person survives, and every item comes with a four-rung hint ladder ending in the exact calculation that decides the outcome.
Unaided, frontier models from Anthropic, Google and OpenAI are right about the person whose fate flips in at most \FrontierDecideHZeroHi\% of items and answer both twins of a pair correctly in at most \FrontierPairsHZeroHi\%.
Pointing at the deciding step helps little, and exact geometry given as text barely helps, yet once the deciding calculation and the quantities it needs, including speeds that a still image cannot show, are written out, the same models reach \FrontierDecideHThreeLo--\FrontierDecideHThreeHi\% on the deciding person, while small open-weight models stay near chance.
Physical prediction from a still image therefore fails at the quantitative step: the strongest models use the deciding quantity when they are given it, but do not produce it on their own.
For models that plan or act in the physical world, a fluent account of a scene is thus no evidence of a correct prediction: such models must find and compute the deciding quantity unprompted, and \dbench{} tests whether they do.
The benchmark, its generator and the evaluation code are available at \url{https://huggingface.co/datasets/emilioferrara/DominoBench}.
\end{abstract}

\section{Introduction}
\label{sec:intro}

Vision-language models have become a common interface between machines and pictures of the physical world: they answer questions about photographs and diagrams, and they are increasingly asked to act in physical environments, where describing a scene well does not translate into predicting it \cite{xu2025deepphy,chow2025physbench}.
Acting rests on an ability that is rarely tested directly, namely predicting what a scene will do next.
A model that sees a ball at the top of a ramp, a ledge, a pendulum and a row of dominoes should be able to say where the ball goes, what it hits, and who is in its way.
Physical reasoning benchmarks have probed this ability for years, through interactive puzzles, videos and textbook questions \cite{bakhtin2019phyre,riochet2022intphys,yi2020clevrer,bear2021physion,chow2025physbench,zhang2025physreason}, and recent benchmarks have begun to ask for forward prediction from a single image \cite{wang2026billiardphys,triebel2026vlatim,mak2026physicsmind}.

\Cref{fig:teaser} shows the kind of scene that exposes the limits of current models.
A pushed ball sets off two rows of dominoes, and a small ball at the end of the chain drops into the bucket of a seesaw whose spikes hang over a person, C, lying on the ground.
The two items of the pair differ in one quantity, the width of the stone counterweight on the other arm of the seesaw.
In item (a) it is 1.25 grid squares, the ball in the bucket outweighs it, and the spikes come down on C. In item (b) it is 1.50 squares and the seesaw does not tip.
Unaided, Claude Opus 5.5 describes a plausible chain reaction and gets both items wrong, predicting that C dies in (b) and survives in (a). Given the calculation that decides the outcome, it gets both right.
The scene is easy to see and the calculation is short. What the model lacks is the step from the picture to the number that decides the outcome.
Existing benchmarks rarely isolate this failure: their items depend on one mechanism, unfold over video, or do not sit close enough to a tipping point for the qualitative chain reaction to be wrong.

This paper asks whether multimodal models can predict the outcome of a multi-step chain reaction from a single still image, and where they fail when they cannot.
To answer it we built \dbench, a generator and benchmark of 2D rigid-body scenes in which a single push starts a chain of three to five mechanisms drawn from seven: ledges with a drop, pits, pendulums, notched balls rolling past a person's head, seesaws with spikes, rows of dominoes and jumps, joined by ramps (\cref{fig:teaser}).
Every scene is built and settled by a physics simulator, which also produces the ground truth.
Every item has a minimal-pair twin in which one quantity, such as the height of a ledge or the width of a pit, differs by one to three grid steps and the fate of one person flips.
Every item is posed along a four-rung hint ladder, from no hint, to a generic nudge, to a pointer at the deciding step, to the exact calculation that decides the outcome, and in a text-only mode that replaces the picture with its exact geometry.
The ladder and the text-only mode probe three abilities that a single accuracy figure conflates: noticing which quantity matters, reading it from the scene, and carrying out the calculation once its inputs are given.
To rank models on one scale we define DominoScore, a weighted geometric mean of chance-corrected accuracy over the mechanisms, in which harder mechanisms weigh more.

We evaluated \NModels{} models from \NFamilies{} families on \dbench-1K, the full benchmark of \NItems{} items in \NPairs{} minimal pairs over \NMech{} deciding mechanisms, and on the medium \dbench-124; we also release \dbench-38, a small, quick-to-run test set (supplementary material).
Our experiments lead to four findings.
(i) Unaided, no model predicts these scenes reliably: the four Claude models that answered all \NItems{} items are right about the full outcome in at most \ExactHZeroHi\% of items, about the person whose fate flips in \DecideHZeroLo--\DecideHZeroHi\%, and about both twins of a pair in at most \PairsHZeroHi\%.
(ii) The failure is quantitative. Pointing a model at the deciding step helps little, while writing out the calculation and its inputs, some of which a still image cannot show, raises accuracy on the deciding person to \DecideHThreeLo--\DecideHThreeHi\%.
(iii) Misreading the picture explains little of the unaided errors: on \dbench-124, with the exact geometry given as text, the Claude 5.5 models gain at most \LadOneFortyTxtImgMaxDiff{} points unaided.
(iv) The pattern holds across providers on \dbench-124, although the two GPT-6 models gain least from the calculation, and the full calculation separates models that can execute it, including most larger open-weight models, from small open-weight models that stay near chance in image mode even when it is written out.

\dbench-1K, with its images, structured scene descriptions, simulator trajectories, hint ladders and twins, is publicly available together with the generator and the evaluation harness at \url{https://huggingface.co/datasets/emilioferrara/DominoBench}.
\section{Related work}
\label{sec:related}

\paragraph{Physical reasoning benchmarks.}
Benchmarks for physical reasoning began with agents and videos \cite{melnik2023benchmarks}: placing objects in a 2D simulated scene until a goal is reached \cite{bakhtin2019phyre}, detecting violations of object permanence \cite{riochet2022intphys,bordes2025intphys2}, and answering predictive, causal and counterfactual questions about collisions, rolling and dominoes in video \cite{yi2020clevrer,bear2021physion,tung2023physionpp,ates2022craft,baradel2020cophy}; video generators still produce physically wrong outcomes \cite{motamed2025physicsiq}.
Predicting dynamics from a single still has a shorter history, from Newtonian scenarios for one object \cite{mottaghi2016newtonian} to the stability of block stacks \cite{groth2018shapestacks,battaglia2013simulation}.
With vision-language models, the field turned to question answering: PhysBench and PhysVista probe properties, relations and dynamics \cite{chow2025physbench,peng2026physvista}, QuantiPhy asks for physical quantities \cite{puyin2025quantiphy}, PhysReason, SeePhys and PHYBench pose multi-step exam problems \cite{zhang2025physreason,xiang2025seephys,qiu2025phybench}, causal benchmarks probe interventions on static scenes \cite{komanduri2025causalvlbench,gao2026ccrv}, and multimodal models fall short of humans on block-tower stability, causal judgments and the conservation of quantities \cite{schulzebuschoff2025visual,luo2026conservation}.
The closest recent work predicts outcomes from one image: billiard trajectories with secondary collisions from the initial frame \cite{wang2026billiardphys}, what a puzzle-game contraption does once started, from a screenshot \cite{triebel2026vlatim}, and the center of mass, lever equilibrium and Newton's first law \cite{mak2026physicsmind}.
We know of no earlier benchmark combining simulated multi-mechanism causal chains in a single still image with tipping-point minimal pairs, a hint ladder and an exact text rendering.

\paragraph{Perception and modality in vision-language models.}
Vision-language models often fail to read exact geometry from an image, such as whether two lines cross or two circles touch \cite{rahmanzadehgervi2024blind,fu2024blink,tong2024eyes}, quantitative visual reasoning in mathematics remains hard \cite{lu2024mathvista}, and models that recognize the content of an image often cannot infer what a transformation of it would show \cite{wang2026rotout}.
Studies that present the same content as an image and as text find that models frequently do better from text \cite{fu2024isobench,wang2024spatialeval,zhang2024mathverse}.
Like these studies, we give every item an exact text rendering, to test whether misreading the picture explains prediction errors.

\paragraph{Minimal pairs.}
Contrast sets and counterfactual edits probe a model's decision boundary with small changes that flip the label \cite{gardner2020contrast,kaushik2020learning}, and paired items with opposite answers expose models that rely on priors in visual question answering \cite{goyal2017making,thrush2022winoground,li2024naturalbench}.
In physical understanding, Minimal Video Pairs scores models on curated video twins with opposite answers \cite{krojer2025mvp}, vision-language models keep the same choice when the deciding measurement in an image pair changes \cite{saha2026newevidence}, and GeoFlip pairs scenes in which one geometric change flips a simulated one-step event for video generators \cite{li2026geoflip}.
We know of no benchmark that builds such twins for causal chains in still images, by stepping one visible parameter across a simulated tipping point.

\paragraph{Hints and the evaluation of reasoning.}
Step-by-step and progressive-hint prompting improve multi-step problem solving \cite{wei2022cot,zheng2023php}, process-level evaluation judges intermediate steps as well as final answers \cite{uesato2022process,lightman2023verify,tang2026causalphys}, PhysicsMind conditions answers on the relevant physical law \cite{mak2026physicsmind}, GRIP locates failures with a ladder of compositional sub-questions \cite{jawaid2026geometricparrots}, and in text-only physics, errors arise more often in building the model of a problem than in solving it \cite{yang2026axiomatic}.
The \dbench{} hint ladder instead adds graded external information, from a generic nudge to a pointer at the deciding step to the deciding calculation itself, so that the rung at which a model starts to succeed indicates what it is missing.
Procedural generation with a simulator follows CLEVR and Kubric \cite{johnson2017clevr,greff2022kubric}, and allows regeneration from fresh seeds, a defense against contamination \cite{mirzadeh2025gsmsymbolic,white2025livebench}.

\section{\dbench}
\label{sec:benchmark}

\dbench{} items are built from physics: a simulator composes a scene, lets it settle, and renders the settled state, so the picture always shows a scene at rest and the ground truth is what the physics does once the ball is pushed.

\paragraph{Scenes.}
A scene is a side view of terrain on a visible grid, with one unit per grid square, populated by people labelled A to F.
Person A stands at the top and pushes a ball, stone or wood, full or with a crescent notch cut out of it.
The ball then passes through a chain of three to five mechanisms drawn from a library of seven, each of which can kill one person: a \emph{ledge}, where the ball flies off a drop and may clear or hit a person standing below; a \emph{pit}, where a person crouches inside a gap the ball may or may not jump; a \emph{pendulum}, a hanging ball that the arriving ball sets swinging toward a person; a \emph{recess}, where a person's head sticks out of a cage and survives only if the ball's notch faces down as it passes; a \emph{seesaw} with downward spikes, which the ball may or may not tip onto a person lying underneath; a row of \emph{dominoes}, which may or may not fall far enough to reach a person or hand off to a second ball; and a \emph{jump}, a kicker that launches the ball over a wall.
Each mechanism is a parameterized part with continuous placement parameters such as the height of a drop, the width of a pit, the length of a pendulum, the gap between dominoes, or the counterweight on a seesaw.
A scene program, stored as plain JSON, lists the parts and their parameters and builds deterministically in pymunk \cite{pymunk}.
The rules of the world are stated with every item: 2D rigid bodies with mass equal to density times area (stone 2.5, wood 0.6 per unit area), balls rolling without slipping except when skidding after a landing, gravity of 9.81 squares/s$^2$, a push that sets the first ball rolling at 1.5 squares/s, and a person who dies when any moving solid object touches them and who neither moves nor stops objects.
The last rule is a convention that keeps every outcome categorical and readable from the trajectory.

\paragraph{Generation checks.}
Every item passes five checks before it enters the benchmark.
The \emph{settle} check runs the scene without the push and rejects it if anything moves more than 0.05 squares or touches a person, and the \emph{trigger} and \emph{chain} checks confirm that the push starts the chain and reaches at least three mechanisms in both twins.
The \emph{stability} check repeats the simulation at half the timestep with twice the solver iterations, with friction scaled by 0.9 and 1.1, and with elasticity raised by 0.05, and rejects the item if any person's fate changes.
The \emph{solvability} check, described below, guarantees that an idealized calculation on the quantities stated in the H3 hint reproduces both outcomes.

\paragraph{Minimal pairs at a tipping point.}
Each item has a twin that differs in one placement parameter of one mechanism, by one to three grid steps (lengths step by 0.25 squares), and in which only the person whom that mechanism can kill changes fate while every other person's fate stays the same.
Twins are found by a guided search.
From one simulation of the base scene we read the quantities that each mechanism's outcome depends on, such as the speed of the ball as it arrives at a ledge or the number of rotations it has completed when it reaches a recess.
With those quantities fixed, an idealized calculation for each mechanism gives the person's fate as a function of the mechanism's own placement parameter: a projectile against a box for the ledge, a projectile against the far rim for the pit, energy and reach for the pendulum, gap against domino height for the dominoes, rotation phase against a simulation-calibrated notch window for the recess, torque about the pivot for the seesaw, and a launch trajectory against the wall for the jump.
The search places twins on either side of the predicted threshold and keeps a pair only if simulation confirms the flip, agrees with the idealized calculation on both twins, leaves every other fate unchanged, and passes the stability check on both.
This is the solvability check: for every kept pair, a short calculation on quantities that can be read from the grid or are stated in the H3 hint, including arrival speeds measured in the simulation, predicts both outcomes, so the H3 hint never points to a wrong answer.
\paragraph{Visible deaths.}
A person's collision box reaches slightly beyond the drawn figure near its corners, so we removed every pair in which a labelled death comes from a body that enters the box without touching the figure (\NExclOneK{} of 500 pairs, mostly at ledges, and \NExclFifty{} of 25 in the small benchmark; supplement).
\dbench-1K keeps \NPairs{} pairs; the medium benchmark shrinks from 140 to \NOneForty{} items and the small one from 50 to \NFifty, hence \dbench-124 and \dbench-38.
The DominoScore weights (\cref{sec:setup}) are computed after this filter.
Of the \NPairs{} pairs, \SepOne{} differ by one grid step, \SepTwo{} by two and \SepThree{} by three.

\paragraph{Items.}
Each item consists of the rendered still, a structured scene description that gives every object's shape, material and position in grid units, the question ``Who dies if A pushes the ball?'', the simulated answer for every person, the full simulator trajectory and contact log, the deciding mechanism and parameter, the identifier of its twin, the hint ladder, and the list of assumptions.
Answers are categorical per person; an item is correct only if every fate is right.

\paragraph{The hint ladder.}
Every item is posed at four rungs, each in a fresh context.
At H0 the model receives the image, the rules and the question, and the instructions already ask it to work through the chain and to check any quantity that could change the outcome.
H1 adds a generic nudge (``Before answering, consider the quantitative details in the picture that could change the outcome'').
H2 points at the deciding step without saying what happens there (for pair 4: ``Consider whether the ball landing in the bucket can outweigh the counterweight above C'').
H3 states the idealized calculation for that step with the quantities it needs, including those that cannot be read from a still image, such as the turning effect with which the empty seesaw holds its arm up (``\ldots{} The empty seesaw (plank, bucket, spikes and stone counterweight together) holds its left arm up with a turning effect of about 1.85. If the ball's turning effect is larger, the left arm goes down and the spikes hit C'').
H3 never states the answer. The model still has to carry out the calculation, compare the result with the person's position, and propagate it through the rest of the chain.
By construction of the solvability check, the H3 calculation predicts the simulated fate of the deciding person on all \NItems{} items.
We compare H0 with H2 for the benefit of locating the deciding step, and H2 with H3 for the benefit of the calculation.

\paragraph{Text-only mode.}
Each item can also be posed without the image, with the structured scene description serialized as text, including exact coordinates.
Text mode removes the need to read positions from pixels, so comparing the two modes tests whether misreading the picture explains errors.

\paragraph{\dbench-1K.}
The full benchmark, \dbench-1K, contains \NPairs{} pairs (\NItems{} items), with \PerMechOneKSecondLo{} to \PerMechOneKHi{} pairs for each of the \NMech{} deciding mechanisms except the ledge, which keeps \PerMechOneKLo{} (\cref{sec:benchmark}, visible deaths).
Chains have three mechanisms in \ChainThree{} items, four in \ChainFour{} and five in \ChainFive{}, and items show \PeopleLo{} to \PeopleHi{} people and kill between \DeathsLo{} and \DeathsHi{} of them.
The naive chain-reaction answer, in which everyone except the pusher dies, is wrong for \NaiveWrong{} of the \NItems{} items.
The generator, the scene programs and the random seeds are released with the benchmark, so new items can be drawn from held-out seeds whenever contamination becomes a concern.
Building \dbench-1K took about 10.7 hours of wall-clock time on a single machine running 14 search processes in parallel (at most about 150 CPU-hours).

\section{Experimental setup}
\label{sec:setup}

\paragraph{Models.}
We evaluated \NModels{} models from \NFamilies{} families.
The closed models are seven Claude models (Fable 5.2, Opus 5.5, Sonnet 5.5, Haiku 5.5, Opus 4.1, Sonnet 4.5 and Haiku 4.5), three OpenAI models served through the Codex command-line interface at high reasoning effort (GPT-5.6-Sol, GPT-6-Astra and GPT-6.1-Sol), Gemini 3.8 Flash, and three Gemini 4 models (Hydrogen, Argon and Carbon) through the Antigravity command-line interface at high effort (Gemini 3.8 Flash too on \dbench-124, and through the Gemini API on \dbench-38).
The open-weight models are Qwen2.5-VL-7B-Instruct, Qwen3-VL-8B-Instruct, Qwen3-VL-8B-Thinking, Qwen3-VL-32B-Instruct, Qwen3-VL-30B-A3B-Instruct, Qwen3.6-35B-A3B and Qwen3.8-27B, Kimi-VL-A3B-Thinking, Mistral-Small-3.2-24B, Muse Glimmer 30B, Gemma 4 31B and Gemma 4 26B-A4B, Nemotron 3.5 Super VL, Nemotron 3 Nano Omni, InternVL3.5-38B and dots-3-note. We call open-weight models of at most 16B parameters small and the others larger.
Locally served open-weight models ran under vLLM \cite{kwon2023vllm} on A100, A40, L40S and RTX PRO 6000 GPUs with temperature 0.6, top-$p$ 0.95, and an answer budget of 8{,}192 tokens, raised to 32{,}768 for thinking models, Gemma 4, Qwen3-VL-30B-A3B and Qwen3-VL-32B, which often ran out of tokens at 8{,}192, and Qwen3-VL-8B-Instruct on \dbench-38. InternVL3.5-38B ran with FP8 weights. The other models ran with their providers' default sampling.

\paragraph{Protocol.}
Each combination of item, input mode and hint rung is one independent call in a fresh context.
The prompt holds the rules of the world, the question, the hints up to the requested rung, and either the image or the structured scene description, and it excludes the answer, the twin and the scene program. Hints identify the deciding step and may state the varied quantity.
Models are asked to reason step by step, to list their assumptions, and to end with a JSON object giving every person's fate and a probability of death.
Command-line clients ran with tools disabled or in a read-only sandbox, from an empty directory holding only a neutrally named copy of the image and with no user configuration or memory, so no model could read the dataset files. Antigravity's web search cannot be disabled; its prompts forbid it, and the audited transcripts show none in the reported answers.
An answer that cannot be parsed, or that runs out of tokens before giving one, is scored as wrong.

\paragraph{Evaluation sets.}
We report main results on two benchmark sizes, based on computational restrictions at the time of submission: calls per item (up to eight), GPU time, rate limits and available GPUs.
\dbench-1K, the full benchmark of \NItems{} items, was answered in image mode at H0 and H3 by four Claude models (Opus 5.5, Sonnet 5.5, Haiku 5.5, Opus 4.1), the three OpenAI models and \NOpenFull{} open-weight models.
\dbench-124, the medium benchmark, holds the \NPairsOneForty{} remaining pairs of a subset of 70 (10 per mechanism) chosen as the pairs with the most completed calls, without regard to correctness. It adds Claude Fable 5.2, Sonnet 4.5 and Haiku 4.5, the three Gemini 4 models, Gemini 3.8 Flash, the three OpenAI models and the slower mid-sized open-weight models. For the four Claude models, exact accuracy on it is within \SubsetGap{} points of \dbench-1K (\cref{tab:main}).
We also release \dbench-38, a small, quick-to-run test set of \NPairsFifty{} separately generated pairs over six mechanisms, with the same checks; its results, including the full ladder for six Claude models, are in the supplement.

\paragraph{Metrics.}
We report three accuracies.
\emph{Exact} accuracy counts an item as correct only if every fate is right.
\emph{Deciding-person} accuracy is computed over the person whose fate flips between the twins. Across complete pairs, a valid prediction that is identical for both twins scores exactly 50\% on this person, including always ``dies'' or always ``survives''.
\emph{Both-twins} accuracy is the fraction of pairs in which both items are exactly right, and any answer that is the same for both twins scores zero on it (intervals: 95\% Wilson, over items).

\paragraph{DominoScore.}
To summarize a model in one number, we combine its deciding-person accuracy over the seven mechanisms, weighting each mechanism by its difficulty.
We use deciding-person accuracy alone because it is the only one of the three whose chance level is the same for every mechanism: the floor of exact accuracy depends on how many people a scene holds and how often they survive, and both-twins accuracy adds little beyond the deciding person while halving the sample to pairs.
For mechanism $m$, let $a_m$ be the deciding-person accuracy and $s_m = \max(0,\, 2a_m - 1)$ the skill above the 50\% scored by any answer that is the same for both twins.
\dbench's score is the weighted geometric mean
\begin{equation}
  \mathrm{DominoScore} = 100 \Big( \prod_{m} (\epsilon + s_m)^{w_m} - \epsilon \Big),\ \epsilon = 0.05,
  \label{eq:dominoscore}
\end{equation}
which is 0 for a model whose accuracy is at or below 50\% on every mechanism and 100 for a model that is always right.
The geometric mean penalizes a model that fails one mechanism outright more than an average would, and $\epsilon$ keeps a single failed mechanism from setting the score to zero.
The weights $w_m$ are proportional to each mechanism's difficulty, measured as the error rate on the deciding person, at H3, of the four Claude models that answered every item, with normalized weights floored at half the uniform weight and then renormalized.
They are frozen and released with the benchmark, so later models are scored on the same scale: \WeightList.
We report DominoScore with a 95\% bootstrap interval that resamples whole twin pairs within each mechanism.
For reference we also score three answers that use no model: the oracle, ``everyone survives'', and the naive chain reaction in which everyone except the pusher dies.

\section{Results}
\label{sec:results}

\begin{table*}[!t]
  \centering
  \caption{Accuracy (\%) in image mode, unaided (H0) and with the calculation (H3), on \dbench-1K and \dbench-124 (\dbench-38: supplement). $n$: items answered, complete runs only. \emph{Both twins}: both items of a pair exactly right. \emph{Deciding person}: the person whose fate flips between the twins. $^\dagger$Open weights. Baselines on \dbench-1K: the naive chain reaction (all but the pusher die) and ``everyone survives'' score \NaiveExact\% and \SurviveExact\% exact, 0\% on both twins and 50\% on the deciding person, an oracle 100\%. Bold: best per column and benchmark; ranked by the shaded column. Data: \dbench-1K and \dbench-124.}
  \label{tab:main}
  \footnotesize
  \renewcommand{\arraystretch}{0.82}
  \begin{tabular}{l r rr rr rr r>{\columncolor{black!9}}r}
\toprule
 & & \multicolumn{2}{c}{Exact} & \multicolumn{2}{c}{Both twins} & \multicolumn{2}{c}{\begin{tabular}[b]{@{}c@{}}Deciding\\person\end{tabular}} & \multicolumn{2}{c}{DominoScore} \\
\cmidrule(lr){3-4}\cmidrule(lr){5-6}\cmidrule(lr){7-8}\cmidrule(lr){9-10}
Model & $n$ & H0 & H3 & H0 & H3 & H0 & H3 & H0 & H3 \\
\midrule
\multicolumn{10}{l}{\textit{\dbench-1K{,} full: all 854 items (427 pairs)}} \\
\midrule
\raisebox{-0.25ex}{\includegraphics[height=1.6ex]{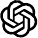}}\,GPT-5.6-Sol & 854 & 42 & 63 & 13 & \textbf{49} & 59 & \textbf{88} & 11 & \textbf{67} \\
\raisebox{-0.25ex}{\includegraphics[height=1.6ex]{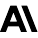}}\,Claude Opus 5.5 & 854 & 47 & 62 & 16 & 44 & 61 & 85 & 12 & 54 \\
\raisebox{-0.25ex}{\includegraphics[height=1.6ex]{figures/logos/pdf/anthropic.pdf}}\,Claude Opus 4.1 & 854 & 45 & 63 & 16 & 45 & 60 & 85 & 12 & 52 \\
\raisebox{-0.25ex}{\includegraphics[height=1.6ex]{figures/logos/pdf/anthropic.pdf}}\,Claude Sonnet 5.5 & 854 & 42 & \textbf{64} & 15 & \textbf{49} & 57 & 86 & 12 & 50 \\
\raisebox{-0.25ex}{\includegraphics[height=1.6ex]{figures/logos/pdf/anthropic.pdf}}\,Claude Haiku 5.5 & 854 & 39 & 53 & 16 & 34 & 59 & 82 & 18 & 47 \\
\raisebox{-0.25ex}{\includegraphics[height=1.6ex]{figures/logos/pdf/openai.pdf}}\,GPT-6.1-Sol & 854 & \textbf{62} & 62 & \textbf{40} & 42 & \textbf{75} & 81 & \textbf{40} & 40 \\
\raisebox{-0.25ex}{\includegraphics[height=1.6ex]{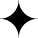}}\,Gemma 4 26B-A4B$^\dagger$ & 854 & 16 & 29 & 1 & 15 & 51 & 76 & 2 & 36 \\
\raisebox{-0.25ex}{\includegraphics[height=1.6ex]{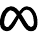}}\,Muse Glimmer 30B$^\dagger$ & 854 & 25 & 43 & 5 & 22 & 53 & 78 & 6 & 34 \\
\raisebox{-0.25ex}{\includegraphics[height=1.6ex]{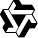}}\,Qwen3.8-27B$^\dagger$ & 854 & 26 & 44 & 8 & 24 & 50 & 79 & 4 & 34 \\
\raisebox{-0.25ex}{\includegraphics[height=1.6ex]{figures/logos/pdf/openai.pdf}}\,GPT-6-Astra & 854 & 60 & 61 & 39 & 42 & \textbf{75} & 80 & 35 & 34 \\
\raisebox{-0.25ex}{\includegraphics[height=1.6ex]{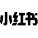}}\,dots-3-note$^\dagger$ & 854 & 24 & 39 & 7 & 20 & 51 & 79 & 3 & 29 \\
\raisebox{-0.25ex}{\includegraphics[height=1.6ex]{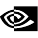}}\,Nemotron 3.5 Super VL$^\dagger$ & 854 & 25 & 44 & 2 & 24 & 51 & 79 & 2 & 26 \\
\rowcolor{black!9}\multicolumn{9}{l}{\footnotesize\textit{4 more models with DominoScore at H3 $\le$ 10: full leaderboard in the appendix}} &  \\
\midrule
\multicolumn{10}{l}{\textit{\dbench-124, medium: subset of \dbench-1K (62 pairs)}} \\
\midrule
\raisebox{-0.25ex}{\includegraphics[height=1.6ex]{figures/logos/pdf/openai.pdf}}\,GPT-5.6-Sol & 124 & 41 & 60 & 16 & 52 & 57 & 90 & 7 & \textbf{75} \\
\raisebox{-0.25ex}{\includegraphics[height=1.6ex]{figures/logos/pdf/googlegemini.pdf}}\,Gemini 4 Hydrogen & 124 & 47 & \textbf{72} & 18 & \textbf{58} & 62 & 90 & 31 & 66 \\
\raisebox{-0.25ex}{\includegraphics[height=1.6ex]{figures/logos/pdf/anthropic.pdf}}\,Claude Opus 5.5 & 124 & 43 & 61 & 13 & 50 & 59 & 91 & 8 & 64 \\
\raisebox{-0.25ex}{\includegraphics[height=1.6ex]{figures/logos/pdf/googlegemini.pdf}}\,Gemini 4 Argon & 124 & 52 & 69 & \textbf{34} & 56 & 71 & \textbf{92} & 40 & 63 \\
\raisebox{-0.25ex}{\includegraphics[height=1.6ex]{figures/logos/pdf/googlegemini.pdf}}\,Gemini 3.8 Flash & 124 & 30 & 61 & 5 & 40 & 52 & 87 & 3 & 63 \\
\raisebox{-0.25ex}{\includegraphics[height=1.6ex]{figures/logos/pdf/googlegemini.pdf}}\,Gemini 4 Carbon & 124 & 52 & 64 & 27 & 50 & 69 & 88 & \textbf{42} & 60 \\
\raisebox{-0.25ex}{\includegraphics[height=1.6ex]{figures/logos/pdf/anthropic.pdf}}\,Claude Fable 5.2 & 124 & 54 & 64 & 24 & 47 & 69 & 85 & 21 & 56 \\
\raisebox{-0.25ex}{\includegraphics[height=1.6ex]{figures/logos/pdf/anthropic.pdf}}\,Claude Opus 4.1 & 124 & 42 & 66 & 10 & 53 & 59 & 88 & 9 & 55 \\
\raisebox{-0.25ex}{\includegraphics[height=1.6ex]{figures/logos/pdf/anthropic.pdf}}\,Claude Haiku 5.5 & 124 & 36 & 52 & 16 & 32 & 57 & 82 & 8 & 48 \\
\raisebox{-0.25ex}{\includegraphics[height=1.6ex]{figures/logos/pdf/anthropic.pdf}}\,Claude Sonnet 5.5 & 124 & 41 & 60 & 18 & 45 & 58 & 86 & 10 & 47 \\
\raisebox{-0.25ex}{\includegraphics[height=1.6ex]{figures/logos/pdf/googlegemini.pdf}}\,Gemma 4 26B-A4B$^\dagger$ & 124 & 16 & 33 & 0 & 21 & 50 & 75 & 3 & 42 \\
\raisebox{-0.25ex}{\includegraphics[height=1.6ex]{figures/logos/pdf/anthropic.pdf}}\,Claude Sonnet 4.5 & 124 & 18 & 40 & 0 & 24 & 48 & 80 & 4 & 40 \\
\raisebox{-0.25ex}{\includegraphics[height=1.6ex]{figures/logos/pdf/openai.pdf}}\,GPT-6-Astra & 124 & 59 & 64 & \textbf{34} & 48 & 70 & 83 & 30 & 36 \\
\raisebox{-0.25ex}{\includegraphics[height=1.6ex]{figures/logos/pdf/openai.pdf}}\,GPT-6.1-Sol & 124 & \textbf{60} & 62 & \textbf{34} & 42 & \textbf{72} & 82 & 25 & 35 \\
\raisebox{-0.25ex}{\includegraphics[height=1.6ex]{figures/logos/pdf/xiaohongshu.pdf}}\,dots-3-note$^\dagger$ & 124 & 29 & 45 & 10 & 21 & 58 & 79 & 20 & 32 \\
\raisebox{-0.25ex}{\includegraphics[height=1.6ex]{figures/logos/pdf/nvidia.pdf}}\,Nemotron 3.5 Super VL$^\dagger$ & 124 & 26 & 46 & 0 & 27 & 52 & 80 & 6 & 28 \\
\raisebox{-0.25ex}{\includegraphics[height=1.6ex]{figures/logos/pdf/meta.pdf}}\,Muse Glimmer 30B$^\dagger$ & 124 & 30 & 47 & 10 & 29 & 55 & 80 & 13 & 24 \\
\rowcolor{black!9}\multicolumn{9}{l}{\footnotesize\textit{13 more models with DominoScore at H3 $\le$ 20: full leaderboard in the appendix}} &  \\
\bottomrule
\end{tabular}

\end{table*}

\begin{figure*}[!t]
  \centering
  \includegraphics[width=\textwidth]{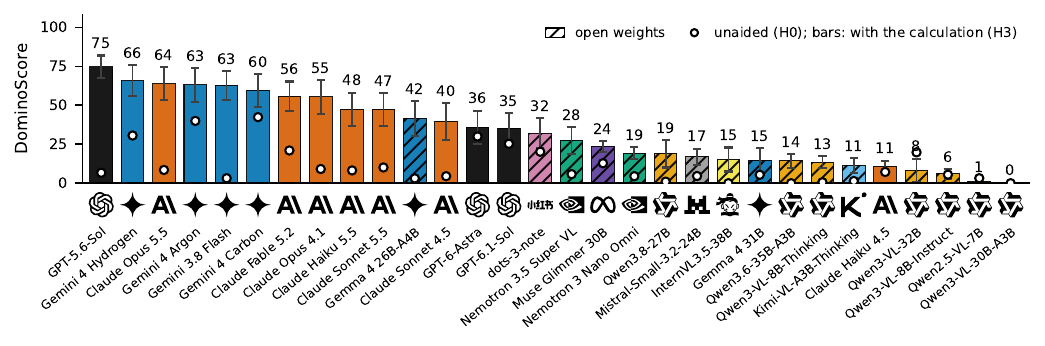}
  \caption{DominoScore of every model with a complete run on \dbench-124, with the deciding calculation (H3, bars with $\pm$1 bootstrap standard error over twin pairs) and unaided (H0, open circles). Neighbouring ranks are often within error. Scores on \dbench-1K are in \cref{tab:main}; the small \dbench-38 is in the appendix.}
  \label{fig:hero}
\end{figure*}

\paragraph{Unaided, no model predicts the outcome reliably.}
The four Claude models that answered all \NItems{} items predict the full outcome (\cref{tab:main}) in \ExactHZeroLo--\ExactHZeroHi\% of items, well above the \NaiveExact\% of the naive chain reaction and the \SurviveExact\% of ``everyone survives''.
That margin comes from the people whose fate does not flip between the twins, whom these models judge correctly \NonFlipHZeroLo--\NonFlipHZeroHi\% of the time.
On the person whose fate flips, the same models are right in \DecideHZeroLo--\DecideHZeroHi\% of items, with intervals that exclude the 50\% of a coin flip by only a few points, and they get both twins of a pair right in \PairsHZeroLo--\PairsHZeroHi\% of pairs.
In \SameAnsHZeroLo--\SameAnsHZeroHi\% of pairs these models give the identical answer to both twins: they describe a plausible chain reaction and do not register that one quantity has moved across a threshold.
With the calculation, the share of identical answers falls to \SameAnsHThreeLo--\SameAnsHThreeHi\%.
Sonnet 4.5 and Haiku 4.5 (48\% and 52\%, \dbench-124) and the open-weight models on \dbench-1K (\OpenOneKHZeroLo--\OpenOneKHZeroHi\%) sit near or below chance on the deciding person and rarely get both twins right.

\paragraph{The pattern holds across model families.}
On \dbench-124, the Gemini models are right about the deciding person in 52--71\% of items unaided and the seven Claude models in 48--69\%; with the calculation they reach 87--92\% and 75--91\% (\cref{tab:main}). GPT-5.6-Sol behaves like them (57\% to 90\%) and ranks first with H3 (DominoScore 75 [58, 87], within error of the next five), on the strength of the pendulum and the jump. GPT-6.1-Sol and GPT-6-Astra do not: they are the most accurate unaided (72\% and 70\% on the deciding person, 60\% and 59\% on the full outcome) but gain least from H3 (82--83\%), so their DominoScore (35--36) trails every Gemini model and all Claude models but Haiku 4.5. On \dbench-1K the two are right unaided in 75\% of items and on both twins in 39--40\% of pairs (Claude: 57--61\% and 15--16\%), yet reach only 80--81\% with H3, and their DominoScore stays flat. GPT-5.6-Sol again goes from 59\% to 88\% and leads \dbench-1K at H3 (DominoScore 67 [61, 73]).

\begin{figure}[!t]
  \centering
  \includegraphics[width=\columnwidth]{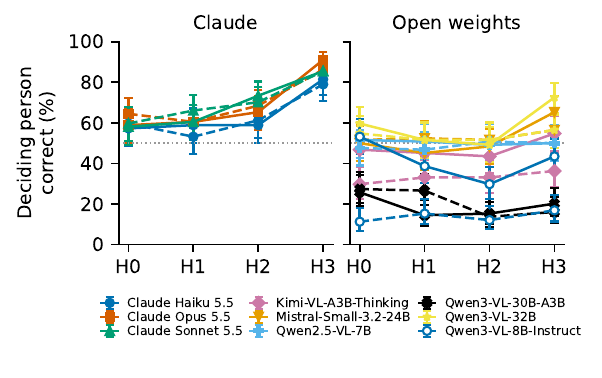}
  \caption{Accuracy on the deciding person along the hint ladder: H0 no hint, H1 a generic nudge, H2 a pointer to the deciding step, H3 the deciding calculation. Left: Claude. Right: open-weight models with a complete ladder. Solid: image mode; dashed: text only. Dotted line: 50\%. Bars: 95\% Wilson intervals. Data: \dbench-124.}
  \label{fig:ladder}
\end{figure}

\paragraph{The calculation helps where a pointer does not.}
On \dbench-124 the three Claude 5.5 models answered every rung in both modes (\cref{fig:ladder}).
H1 adds at most \LadOneFortyHOneMaxGain{} points, as H0 already asks models to check quantities that could change the outcome.
The pointer of H2 helps Claude Sonnet 5.5 in image mode (58\% to 73\%) but changes little for Opus 5.5 (59\% to 65\%) and Haiku 5.5 (57\% to 59\%), while H3 lifts all three, in both modes, from \LadOneFortyHZeroLo--\LadOneFortyHZeroHi\% to \LadOneFortyHThreeLo--\LadOneFortyHThreeHi\%.
On \dbench-1K the effect is the same at scale: the four Claude models that answered all items rise from \DecideHZeroLo--\DecideHZeroHi\% to \DecideHThreeLo--\DecideHThreeHi\% on the deciding person and from \PairsHZeroLo--\PairsHZeroHi\% to \PairsHThreeLo--\PairsHThreeHi\% on both twins.
Knowing where to look is therefore not enough. What the strongest models lack is the step from the location of the decision to a number that settles it.

\paragraph{Coordinates given as text barely improve unaided accuracy.}
On \dbench-124, the three Claude 5.5 models are right about the deciding person in 60--64\% of items from text and 57--59\% from images at H0, and in 79--86\% and 82--91\% at H3.
The two modes differ by at most \LadOneFortyTxtImgMaxDiff{} points at H0, so misreading the picture explains little of the errors.

\paragraph{Small open-weight models gain little even from the calculation.}
On \dbench-1K, the small open-weight models (Qwen2.5-VL-7B, Qwen3-VL-8B-Instruct and Kimi-VL-A3B-Thinking) and Qwen3-VL-30B-A3B gain little or nothing from H3, while Muse Glimmer 30B, Gemma 4 26B-A4B, Qwen3.8-27B, dots-3-note and Nemotron 3.5 Super VL rise from \GainOneKHZeroLo--\GainOneKHZeroHi\% to \GainOneKHThreeLo--\GainOneKHThreeHi\% on the deciding person (DominoScore \GainOneKDSLo--\GainOneKDSHi).
The newest larger ones reach the Claude range on \dbench-124: Qwen3.8-27B and Muse Glimmer 30B rise from \QwenThreeEightHZero\% and \MuseHZero\% to \QwenThreeEightHThree\% and \MuseHThree\% on the deciding person, level with Claude Haiku 5.5 on the same items (\HaikuSubHThree\%).
Qwen3-VL-8B-Instruct and Qwen3-VL-30B-A3B fall below chance by looping past the token budget (\UnparsedQEightInstHThree\% and \UnparsedQThirtyAThreeHThree\% unfinished).

\paragraph{The other open-weight models on \dbench-124.}
The open-weight models that ran only on \dbench-124 stay below the larger ones even with the calculation: Nemotron 3 Nano Omni, Gemma 4 31B, InternVL3.5-38B and Qwen3.6-35B-A3B reach 66--69\% on the deciding person at H3 (DominoScore 14--19), and Qwen3-VL-8B-Thinking 56\%, with 23\% of its H3 answers unfinished.
The six open-weight models with a complete ladder (\cref{fig:ladder}) gain nothing from H1 or H2: Mistral-Small-3.2-24B, Kimi-VL-A3B-Thinking, Qwen2.5-VL-7B, Qwen3-VL-32B and Qwen3-VL-30B-A3B stay level or fall, and Qwen3-VL-8B-Instruct falls from 53\% at H0 to 30\% at H2.
With H3 only Mistral-Small-3.2-24B and Qwen3-VL-32B rise clearly (to 65\% and 73\%), and from text none of the six exceeds 56\%.

\begin{figure}[!t]
  \centering
  \includegraphics[width=\columnwidth]{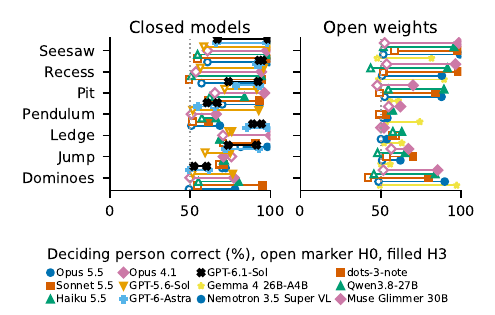}
  \caption{Accuracy on the deciding person by deciding mechanism, from H0 (open) to H3 (filled), image mode, all \NItems{} items. Left: closed models, right: open-weight models above DominoScore 20 at H3. Data: \dbench-1K.}
  \label{fig:mechanism}
\end{figure}

\paragraph{Mechanisms.}
Unaided, the Claude models are near chance (\cref{fig:mechanism}) on the recess, the dominoes and the pendulum (49--57\%), where the outcome depends on a rotation phase, a gap measured against a height or the reach of a swing, and they do best on the ledge, the pit and the jump (62--82\%).
With the calculation, the recess, the seesaw and the pit rise to 84--100\%, the dominoes to 77--95\%, while the pendulum gains least among the mechanisms that start near chance (51--57\% to 61--68\%) and the jump does not move (68--76\% to 71--72\%).
On both, the written-out calculation branches (a strike then a swing; a flight that clears or meets a wall), and the Claude models often fail it.
With H3, the five larger open-weight models of \cref{fig:mechanism} rise well above chance on the dominoes, the recess and the seesaw.

\begin{figure}[!t]
  \centering
  \includegraphics[width=\columnwidth]{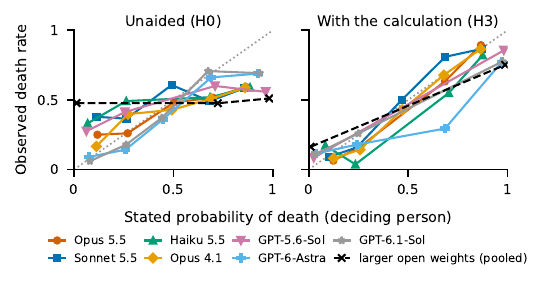}
  \caption{Stated probability of death for the deciding person against how often that person dies, in bins of 0.2 (at least 15 statements each), image mode. Solid: closed models that answered every item; dashed: the five larger open-weight models of \cref{fig:mechanism}, pooled; dotted: perfect calibration. Data: \dbench-1K.}
  \label{fig:calibration}
\end{figure}

\paragraph{Confidence.}
Unaided, the models are confident without being right (\cref{fig:calibration}): the four Claude models that answered all items give the deciding person a death probability of at least 0.8 or at most 0.2 in \CalibConfShareHZeroLo--\CalibConfShareHZeroHi\% of items, and these confident statements are correct in only \CalibConfAccHZeroLo--\CalibConfAccHZeroHi\% (GPT models \CalibGPTConfAccHZeroLo--\CalibGPTConfAccHZeroHi\%, pooled larger open-weight models \CalibOpenConfAccHZero\%).
With H3, their confident statements are correct in \CalibConfAccHThreeLo--\CalibConfAccHThreeHi\%.

\paragraph{DominoScore.}
DominoScore (\cref{eq:dominoscore}) folds these per-mechanism results into one number per model (\cref{tab:main}), and we rank every model with a complete run on the same \NOneForty{} items of \dbench-124 (\cref{fig:hero}).
There, the three Gemini 4 models score highest unaided (\GFourOneFortyHZeroLo--\GFourOneFortyHZeroHi; Claude \ClaudeOneFortyHZeroLo--\ClaudeOneFortyHZeroHi) and reach \GFourOneFortyHThreeLo--\GFourOneFortyHThreeHi{} with the calculation, level with Claude Opus 5.5 (\OpusOneFortyDSHThree).
Unaided, the four Claude models that answered all \NItems{} items score \DSClaudeHZeroLo--\DSClaudeHZeroHi{} and the small open-weight models at most \DSSmallHZeroHi.
With the calculation, the Claude models reach \DSClaudeHThreeLo--\DSClaudeHThreeHi{} (Opus 5.5 highest, \DSOpusHThree{} [\DSOpusHThreeLo, \DSOpusHThreeHi]) and the small open-weight models at most \DSSmallHThreeHi.
\paragraph{Robustness of DominoScore.}
Coin-flip guessers stay below the \cref{tab:main} thresholds (95th percentiles \RandDSOneKPct{} and \RandDSOneFortyPct), equal weights barely change the rankings ($\rho = \UniformRhoOneK$ and \UniformRhoOneForty), and a repeat Opus 5.5 run stays in its interval.
\section{Limitations}
\label{sec:limitations}

\dbench{} scenes are two-dimensional and drawn in a single line-art style, so the benchmark does not test perception of photographs or depth.
The rule that a person dies when touched by any moving solid is a convention that keeps outcomes categorical; it is stated with every item.
Its collision boxes reach slightly beyond the drawn figures; we removed the pairs in which this decides a death (\cref{sec:benchmark}).
H3 supplies quantities that a still image cannot show, such as speeds and thresholds calibrated in simulation, so the top rung measures the execution of a stated calculation. Unaided, a model must estimate those quantities from the rules and the picture, and we did not measure how well they do so.
Coverage is uneven across models (\cref{sec:setup}), and the text-only comparison for Claude rests on \dbench-124.
We did not measure human accuracy; in video prediction, people surpass vision models on similar scenes \cite{bear2021physion,li2026humanlevel}.

\section{Conclusion}
\label{sec:conclusion}

Unaided, models do not predict these chain reactions from a still image: on \dbench-1K, the four Claude models that answered all items are only modestly better than a coin flip on the person whose fate flips, and they usually answer both twins of a pair alike.
We locate the failure in the quantitative step: on \dbench-124 neither a pointer to the deciding step nor exact coordinates help the Claude models much, while on \dbench-1K the written-out calculation brings the four Claude models to \DecideHThreeLo--\DecideHThreeHi\% on that person.
Closing this gap requires models that decide on their own which quantity a scene turns on and then compute it.
\dbench{} tests for that ability, and its generator can produce fresh scenes as models improve.

\section*{LLM Usage Statement}

We used three LLM-based assistants in this project, under the authors' direction and review: Claude Opus 5.5 (Anthropic), through the Claude Code CLI agent, GPT-6-Astra (OpenAI), through the Codex CLI agent, and Claude Fable 5.2 (Anthropic), for a review of the draft.

\paragraph{Ideation.}
The idea for the benchmark is the authors'.
It came from posing a chain-reaction puzzle to Claude Opus 5.5 and GPT-6-Astra and observing that both models missed the quantities that decided the outcome until they were pointed out.
We then used Claude Opus 5.5 as a discussion partner while refining the design (minimal pairs, the hint ladder, the text-only mode).

\paragraph{Code and experiments.}
Claude Opus 5.5 wrote most of the code for the scene generator, the physics-simulation and labelling pipeline, the evaluation harness, the analysis scripts and the figure scripts, and it launched and monitored the evaluation runs.
The authors specified the design, reviewed the code and checked results against the simulator.
Model-performance statistics are computed from stored responses. Dataset statistics and physical quantities come from the generated scenes and simulations.

\paragraph{Related work.}
Claude Opus 5.5, through the Claude Code CLI agent, searched the literature and proposed references. Every reference was checked against Crossref or the arXiv record before inclusion.

\paragraph{Text.}
Claude Opus 5.5 and GPT-6-Astra drafted and reviewed the text of the paper from the authors' instructions. The authors reviewed, revised, edited, and approved it.

\paragraph{Models as subjects.}
Claude and GPT models are also among the models evaluated.
They were queried in isolated sessions, the Claude models with tools disabled and the GPT models in a read-only sandbox, each from a temporary directory holding at most a copy of the item's image, with no access to the project's files, notes or prior conversation, and answered the same prompts as every other model.

{\small
\bibliographystyle{ieeenat_fullname}
\bibliography{refs}
}
\clearpage
\onecolumn
\appendix

\renewcommand{\thesection}{\Alph{section}}
\renewcommand{\thetable}{S\arabic{table}}
\renewcommand{\thefigure}{S\arabic{figure}}
\renewcommand{\theequation}{S\arabic{equation}}

This appendix gives the full leaderboard of each benchmark (\cref{sec:leaderboards}), the full results along the hint ladder (\cref{sec:full}), how often each model answers both twins alike (\cref{sec:same}), the results for each mechanism (\cref{sec:mech}), the rule and the idealized calculation behind each mechanism (\cref{sec:rules}), the pairs removed because a death is not visible in the picture (\cref{sec:visible}), the composition of the benchmarks (\cref{sec:stats}), one example pair per mechanism (\cref{sec:examples}), and the exact prompts, the answer parser and the DominoScore weights (\cref{sec:prompts}).
Every table and figure here is generated by script from the same result files as the main text.

\FloatBarrier
\section{Full leaderboards}
\label{sec:leaderboards}

\begin{figure*}[!htbp]
  \centering
  \includegraphics[width=\textwidth]{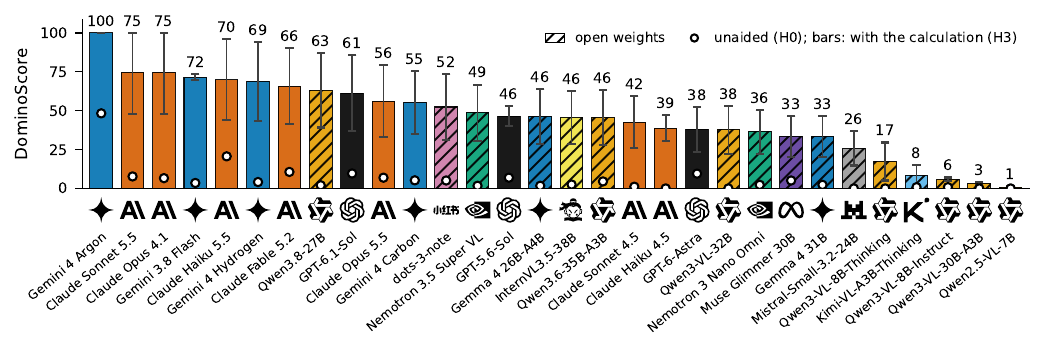}
  \caption{DominoScore of every model on \dbench-38, the one benchmark every model answered, with the deciding calculation (H3, bars with $\pm$1 bootstrap standard error over twin pairs) and unaided (H0, open circles). Only complete runs are shown. Scores rest on six mechanisms with the frozen weights renormalized, so neighbouring ranks are often within error. \Cref{fig:hero} gives the same ranking on \dbench-124.}
  \label{fig:supp-hero-50}
\end{figure*}

\dbench-38 is a small, quick-to-run test set: the main text reports \dbench-1K and \dbench-124, and \dbench-38 is reported here only.
\Cref{tab:main} lists the models whose DominoScore at H3, as printed (rounded), exceeds 10 on \dbench-1K and 20 on \dbench-124.
\Cref{fig:supp-hero-50} ranks every model on \dbench-38, the one benchmark that all models answered; \cref{fig:hero} ranks them on \dbench-124.
\Cref{tab:supp-lb-1k,tab:supp-lb-140,tab:supp-lb-50} give every model with a complete run on each benchmark, with the same columns and ranked the same way.
Which models ran on which benchmark was set by cost (\cref{sec:setup}); the models with a complete run (image mode, H0 and H3 on every item) on each benchmark are listed below.

\paragraph{Run on \dbench-1K (16 models).} Claude Opus 5.5, Claude Sonnet 5.5, Claude Haiku 5.5, Claude Opus 4.1, GPT-5.6-Sol, GPT-6-Astra, GPT-6.1-Sol, Gemma 4 26B-A4B, Nemotron 3.5 Super VL, dots-3-note, Qwen2.5-VL-7B, Qwen3-VL-8B-Instruct, Qwen3-VL-30B-A3B, Qwen3.8-27B, Muse Glimmer 30B, Kimi-VL-A3B-Thinking.
\paragraph{Run on \dbench-124 (30 models).} Claude Opus 5.5, Claude Sonnet 5.5, Claude Haiku 5.5, Claude Opus 4.1, Claude Sonnet 4.5, Claude Haiku 4.5, Claude Fable 5.2, GPT-5.6-Sol, GPT-6-Astra, GPT-6.1-Sol, Gemini 4 Hydrogen, Gemini 4 Carbon, Gemini 4 Argon, Gemini 3.8 Flash, Gemma 4 31B, Gemma 4 26B-A4B, Nemotron 3.5 Super VL, Nemotron 3 Nano Omni, dots-3-note, Qwen2.5-VL-7B, Qwen3-VL-8B-Instruct, Qwen3-VL-8B-Thinking, Qwen3-VL-32B, Qwen3-VL-30B-A3B, Qwen3.8-27B, Qwen3.6-35B-A3B, Muse Glimmer 30B, InternVL3.5-38B, Mistral-Small-3.2-24B, Kimi-VL-A3B-Thinking.
\paragraph{Run on \dbench-38 (30 models).} Claude Opus 5.5, Claude Sonnet 5.5, Claude Haiku 5.5, Claude Opus 4.1, Claude Sonnet 4.5, Claude Haiku 4.5, Claude Fable 5.2, GPT-5.6-Sol, GPT-6-Astra, GPT-6.1-Sol, Gemini 3.8 Flash, Gemini 4 Hydrogen, Gemini 4 Carbon, Gemini 4 Argon, Gemma 4 31B, Gemma 4 26B-A4B, Nemotron 3.5 Super VL, Nemotron 3 Nano Omni, dots-3-note, Qwen2.5-VL-7B, Qwen3-VL-8B-Instruct, Qwen3-VL-8B-Thinking, Qwen3-VL-32B, Qwen3-VL-30B-A3B, Qwen3.8-27B, Qwen3.6-35B-A3B, Muse Glimmer 30B, InternVL3.5-38B, Mistral-Small-3.2-24B, Kimi-VL-A3B-Thinking.

\begin{table*}[!htbp]
  \centering
  \caption{Full leaderboard on \dbench-1K (all \NItems{} items, \NPairs{} pairs), image mode, unaided (H0) and with the calculation (H3). Columns as in \cref{tab:main}. $^\dagger$Open weights. Bold: best in each column. Ranked by DominoScore at H3. Data: \dbench-1K.}
  \label{tab:supp-lb-1k}
  \footnotesize
  \begin{tabular}{l r rr rr rr r>{\columncolor{black!9}}r}
\toprule
 & & \multicolumn{2}{c}{Exact} & \multicolumn{2}{c}{Both twins} & \multicolumn{2}{c}{\begin{tabular}[b]{@{}c@{}}Deciding\\person\end{tabular}} & \multicolumn{2}{c}{DominoScore} \\
\cmidrule(lr){3-4}\cmidrule(lr){5-6}\cmidrule(lr){7-8}\cmidrule(lr){9-10}
Model & $n$ & H0 & H3 & H0 & H3 & H0 & H3 & H0 & H3 \\
\midrule
\raisebox{-0.25ex}{\includegraphics[height=1.6ex]{figures/logos/pdf/openai.pdf}}\,GPT-5.6-Sol & 854 & 42 & 63 & 13 & \textbf{49} & 59 & \textbf{88} & 11 & \textbf{67} \\
\raisebox{-0.25ex}{\includegraphics[height=1.6ex]{figures/logos/pdf/anthropic.pdf}}\,Claude Opus 5.5 & 854 & 47 & 62 & 16 & 44 & 61 & 85 & 12 & 54 \\
\raisebox{-0.25ex}{\includegraphics[height=1.6ex]{figures/logos/pdf/anthropic.pdf}}\,Claude Opus 4.1 & 854 & 45 & 63 & 16 & 45 & 60 & 85 & 12 & 52 \\
\raisebox{-0.25ex}{\includegraphics[height=1.6ex]{figures/logos/pdf/anthropic.pdf}}\,Claude Sonnet 5.5 & 854 & 42 & \textbf{64} & 15 & \textbf{49} & 57 & 86 & 12 & 50 \\
\raisebox{-0.25ex}{\includegraphics[height=1.6ex]{figures/logos/pdf/anthropic.pdf}}\,Claude Haiku 5.5 & 854 & 39 & 53 & 16 & 34 & 59 & 82 & 18 & 47 \\
\raisebox{-0.25ex}{\includegraphics[height=1.6ex]{figures/logos/pdf/openai.pdf}}\,GPT-6.1-Sol & 854 & \textbf{62} & 62 & \textbf{40} & 42 & \textbf{75} & 81 & \textbf{40} & 40 \\
\raisebox{-0.25ex}{\includegraphics[height=1.6ex]{figures/logos/pdf/googlegemini.pdf}}\,Gemma 4 26B-A4B$^\dagger$ & 854 & 16 & 29 & 1 & 15 & 51 & 76 & 2 & 36 \\
\raisebox{-0.25ex}{\includegraphics[height=1.6ex]{figures/logos/pdf/meta.pdf}}\,Muse Glimmer 30B$^\dagger$ & 854 & 25 & 43 & 5 & 22 & 53 & 78 & 6 & 34 \\
\raisebox{-0.25ex}{\includegraphics[height=1.6ex]{figures/logos/pdf/qwen.pdf}}\,Qwen3.8-27B$^\dagger$ & 854 & 26 & 44 & 8 & 24 & 50 & 79 & 4 & 34 \\
\raisebox{-0.25ex}{\includegraphics[height=1.6ex]{figures/logos/pdf/openai.pdf}}\,GPT-6-Astra & 854 & 60 & 61 & 39 & 42 & \textbf{75} & 80 & 35 & 34 \\
\raisebox{-0.25ex}{\includegraphics[height=1.6ex]{figures/logos/pdf/xiaohongshu.pdf}}\,dots-3-note$^\dagger$ & 854 & 24 & 39 & 7 & 20 & 51 & 79 & 3 & 29 \\
\raisebox{-0.25ex}{\includegraphics[height=1.6ex]{figures/logos/pdf/nvidia.pdf}}\,Nemotron 3.5 Super VL$^\dagger$ & 854 & 25 & 44 & 2 & 24 & 51 & 79 & 2 & 26 \\
\raisebox{-0.25ex}{\includegraphics[height=1.6ex]{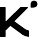}}\,Kimi-VL-A3B-Thinking$^\dagger$ & 854 & 21 & 25 & 3 & 7 & 50 & 56 & 2 & 10 \\
\raisebox{-0.25ex}{\includegraphics[height=1.6ex]{figures/logos/pdf/qwen.pdf}}\,Qwen2.5-VL-7B$^\dagger$ & 854 & 20 & 24 & 4 & 8 & 47 & 54 & 1 & 8 \\
\raisebox{-0.25ex}{\includegraphics[height=1.6ex]{figures/logos/pdf/qwen.pdf}}\,Qwen3-VL-8B-Instruct$^\dagger$ & 854 & 16 & 16 & 2 & 5 & 46 & 44 & 0 & 5 \\
\raisebox{-0.25ex}{\includegraphics[height=1.6ex]{figures/logos/pdf/qwen.pdf}}\,Qwen3-VL-30B-A3B$^\dagger$ & 854 & 10 & 10 & 0 & 3 & 25 & 21 & 0 & 0 \\
\bottomrule
\end{tabular}

\end{table*}

\begin{table*}[!htbp]
  \centering
  \caption{Full leaderboard on \dbench-124 (subset of \dbench-1K, \NPairsOneForty{} pairs), image mode, H0 and H3. Columns as in \cref{tab:main}. $^\dagger$Open weights. Bold: best in each column. Ranked by DominoScore at H3. Data: \dbench-124.}
  \label{tab:supp-lb-140}
  \footnotesize
  \begin{tabular}{l r rr rr rr r>{\columncolor{black!9}}r}
\toprule
 & & \multicolumn{2}{c}{Exact} & \multicolumn{2}{c}{Both twins} & \multicolumn{2}{c}{\begin{tabular}[b]{@{}c@{}}Deciding\\person\end{tabular}} & \multicolumn{2}{c}{DominoScore} \\
\cmidrule(lr){3-4}\cmidrule(lr){5-6}\cmidrule(lr){7-8}\cmidrule(lr){9-10}
Model & $n$ & H0 & H3 & H0 & H3 & H0 & H3 & H0 & H3 \\
\midrule
\raisebox{-0.25ex}{\includegraphics[height=1.6ex]{figures/logos/pdf/openai.pdf}}\,GPT-5.6-Sol & 124 & 41 & 60 & 16 & 52 & 57 & 90 & 7 & \textbf{75} \\
\raisebox{-0.25ex}{\includegraphics[height=1.6ex]{figures/logos/pdf/googlegemini.pdf}}\,Gemini 4 Hydrogen & 124 & 47 & \textbf{72} & 18 & \textbf{58} & 62 & 90 & 31 & 66 \\
\raisebox{-0.25ex}{\includegraphics[height=1.6ex]{figures/logos/pdf/anthropic.pdf}}\,Claude Opus 5.5 & 124 & 43 & 61 & 13 & 50 & 59 & 91 & 8 & 64 \\
\raisebox{-0.25ex}{\includegraphics[height=1.6ex]{figures/logos/pdf/googlegemini.pdf}}\,Gemini 4 Argon & 124 & 52 & 69 & \textbf{34} & 56 & 71 & \textbf{92} & 40 & 63 \\
\raisebox{-0.25ex}{\includegraphics[height=1.6ex]{figures/logos/pdf/googlegemini.pdf}}\,Gemini 3.8 Flash & 124 & 30 & 61 & 5 & 40 & 52 & 87 & 3 & 63 \\
\raisebox{-0.25ex}{\includegraphics[height=1.6ex]{figures/logos/pdf/googlegemini.pdf}}\,Gemini 4 Carbon & 124 & 52 & 64 & 27 & 50 & 69 & 88 & \textbf{42} & 60 \\
\raisebox{-0.25ex}{\includegraphics[height=1.6ex]{figures/logos/pdf/anthropic.pdf}}\,Claude Fable 5.2 & 124 & 54 & 64 & 24 & 47 & 69 & 85 & 21 & 56 \\
\raisebox{-0.25ex}{\includegraphics[height=1.6ex]{figures/logos/pdf/anthropic.pdf}}\,Claude Opus 4.1 & 124 & 42 & 66 & 10 & 53 & 59 & 88 & 9 & 55 \\
\raisebox{-0.25ex}{\includegraphics[height=1.6ex]{figures/logos/pdf/anthropic.pdf}}\,Claude Haiku 5.5 & 124 & 36 & 52 & 16 & 32 & 57 & 82 & 8 & 48 \\
\raisebox{-0.25ex}{\includegraphics[height=1.6ex]{figures/logos/pdf/anthropic.pdf}}\,Claude Sonnet 5.5 & 124 & 41 & 60 & 18 & 45 & 58 & 86 & 10 & 47 \\
\raisebox{-0.25ex}{\includegraphics[height=1.6ex]{figures/logos/pdf/googlegemini.pdf}}\,Gemma 4 26B-A4B$^\dagger$ & 124 & 16 & 33 & 0 & 21 & 50 & 75 & 3 & 42 \\
\raisebox{-0.25ex}{\includegraphics[height=1.6ex]{figures/logos/pdf/anthropic.pdf}}\,Claude Sonnet 4.5 & 124 & 18 & 40 & 0 & 24 & 48 & 80 & 4 & 40 \\
\raisebox{-0.25ex}{\includegraphics[height=1.6ex]{figures/logos/pdf/openai.pdf}}\,GPT-6-Astra & 124 & 59 & 64 & \textbf{34} & 48 & 70 & 83 & 30 & 36 \\
\raisebox{-0.25ex}{\includegraphics[height=1.6ex]{figures/logos/pdf/openai.pdf}}\,GPT-6.1-Sol & 124 & \textbf{60} & 62 & \textbf{34} & 42 & \textbf{72} & 82 & 25 & 35 \\
\raisebox{-0.25ex}{\includegraphics[height=1.6ex]{figures/logos/pdf/xiaohongshu.pdf}}\,dots-3-note$^\dagger$ & 124 & 29 & 45 & 10 & 21 & 58 & 79 & 20 & 32 \\
\raisebox{-0.25ex}{\includegraphics[height=1.6ex]{figures/logos/pdf/nvidia.pdf}}\,Nemotron 3.5 Super VL$^\dagger$ & 124 & 26 & 46 & 0 & 27 & 52 & 80 & 6 & 28 \\
\raisebox{-0.25ex}{\includegraphics[height=1.6ex]{figures/logos/pdf/meta.pdf}}\,Muse Glimmer 30B$^\dagger$ & 124 & 30 & 47 & 10 & 29 & 55 & 80 & 13 & 24 \\
\raisebox{-0.25ex}{\includegraphics[height=1.6ex]{figures/logos/pdf/nvidia.pdf}}\,Nemotron 3 Nano Omni$^\dagger$ & 124 & 20 & 30 & 0 & 5 & 54 & 67 & 4 & 19 \\
\raisebox{-0.25ex}{\includegraphics[height=1.6ex]{figures/logos/pdf/qwen.pdf}}\,Qwen3.8-27B$^\dagger$ & 124 & 24 & 47 & 6 & 26 & 47 & 82 & 1 & 19 \\
\raisebox{-0.25ex}{\includegraphics[height=1.6ex]{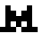}}\,Mistral-Small-3.2-24B$^\dagger$ & 124 & 17 & 23 & 2 & 6 & 50 & 65 & 5 & 17 \\
\raisebox{-0.25ex}{\includegraphics[height=1.6ex]{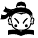}}\,InternVL3.5-38B$^\dagger$ & 124 & 12 & 23 & 3 & 10 & 45 & 69 & 0 & 15 \\
\raisebox{-0.25ex}{\includegraphics[height=1.6ex]{figures/logos/pdf/googlegemini.pdf}}\,Gemma 4 31B$^\dagger$ & 124 & 22 & 32 & 0 & 8 & 53 & 69 & 5 & 15 \\
\raisebox{-0.25ex}{\includegraphics[height=1.6ex]{figures/logos/pdf/qwen.pdf}}\,Qwen3.6-35B-A3B$^\dagger$ & 124 & 18 & 33 & 0 & 10 & 44 & 66 & 0 & 14 \\
\raisebox{-0.25ex}{\includegraphics[height=1.6ex]{figures/logos/pdf/qwen.pdf}}\,Qwen3-VL-8B-Thinking$^\dagger$ & 124 & 18 & 18 & 2 & 8 & 45 & 56 & 1 & 13 \\
\raisebox{-0.25ex}{\includegraphics[height=1.6ex]{figures/logos/pdf/kimi.pdf}}\,Kimi-VL-A3B-Thinking$^\dagger$ & 124 & 19 & 22 & 5 & 8 & 47 & 55 & 1 & 11 \\
\raisebox{-0.25ex}{\includegraphics[height=1.6ex]{figures/logos/pdf/anthropic.pdf}}\,Claude Haiku 4.5 & 124 & 15 & 25 & 2 & 10 & 52 & 75 & 7 & 11 \\
\raisebox{-0.25ex}{\includegraphics[height=1.6ex]{figures/logos/pdf/qwen.pdf}}\,Qwen3-VL-32B$^\dagger$ & 124 & 23 & 26 & 5 & 11 & 60 & 73 & 20 & 8 \\
\raisebox{-0.25ex}{\includegraphics[height=1.6ex]{figures/logos/pdf/qwen.pdf}}\,Qwen3-VL-8B-Instruct$^\dagger$ & 124 & 18 & 14 & 2 & 3 & 53 & 44 & 6 & 6 \\
\raisebox{-0.25ex}{\includegraphics[height=1.6ex]{figures/logos/pdf/qwen.pdf}}\,Qwen2.5-VL-7B$^\dagger$ & 124 & 22 & 24 & 3 & 8 & 52 & 50 & 3 & 1 \\
\raisebox{-0.25ex}{\includegraphics[height=1.6ex]{figures/logos/pdf/qwen.pdf}}\,Qwen3-VL-30B-A3B$^\dagger$ & 124 & 13 & 10 & 2 & 3 & 26 & 20 & 0 & 0 \\
\bottomrule
\end{tabular}

\end{table*}

\begin{table*}[!htbp]
  \centering
  \caption{Full leaderboard on \dbench-38 (\NPairsFifty{} pairs, six mechanisms), image mode, H0 and H3. Columns as in \cref{tab:main}. DominoScore uses the frozen weights renormalized over the six mechanisms. $^\dagger$Open weights. Bold: best in each column. Ranked by DominoScore at H3. Data: \dbench-38.}
  \label{tab:supp-lb-50}
  \footnotesize
  \begin{tabular}{l r rr rr rr r>{\columncolor{black!9}}r}
\toprule
 & & \multicolumn{2}{c}{Exact} & \multicolumn{2}{c}{Both twins} & \multicolumn{2}{c}{\begin{tabular}[b]{@{}c@{}}Deciding\\person\end{tabular}} & \multicolumn{2}{c}{DominoScore} \\
\cmidrule(lr){3-4}\cmidrule(lr){5-6}\cmidrule(lr){7-8}\cmidrule(lr){9-10}
Model & $n$ & H0 & H3 & H0 & H3 & H0 & H3 & H0 & H3 \\
\midrule
\raisebox{-0.25ex}{\includegraphics[height=1.6ex]{figures/logos/pdf/googlegemini.pdf}}\,Gemini 4 Argon & 38 & 53 & \textbf{79} & \textbf{32} & \textbf{74} & 71 & \textbf{100} & \textbf{48} & \textbf{100} \\
\raisebox{-0.25ex}{\includegraphics[height=1.6ex]{figures/logos/pdf/anthropic.pdf}}\,Claude Sonnet 5.5 & 38 & 42 & 55 & 21 & 47 & 58 & 95 & 8 & 75 \\
\raisebox{-0.25ex}{\includegraphics[height=1.6ex]{figures/logos/pdf/anthropic.pdf}}\,Claude Opus 4.1 & 38 & 42 & 60 & 26 & 53 & 66 & 95 & 7 & 75 \\
\raisebox{-0.25ex}{\includegraphics[height=1.6ex]{figures/logos/pdf/googlegemini.pdf}}\,Gemini 3.8 Flash & 38 & 29 & 58 & 10 & 47 & 53 & 92 & 4 & 72 \\
\raisebox{-0.25ex}{\includegraphics[height=1.6ex]{figures/logos/pdf/anthropic.pdf}}\,Claude Haiku 5.5 & 38 & 34 & 53 & 5 & 37 & 63 & 95 & 21 & 70 \\
\raisebox{-0.25ex}{\includegraphics[height=1.6ex]{figures/logos/pdf/googlegemini.pdf}}\,Gemini 4 Hydrogen & 38 & 47 & 63 & 16 & 47 & 55 & 92 & 4 & 69 \\
\raisebox{-0.25ex}{\includegraphics[height=1.6ex]{figures/logos/pdf/anthropic.pdf}}\,Claude Fable 5.2 & 38 & 50 & 55 & \textbf{32} & 37 & \textbf{79} & 87 & 11 & 66 \\
\raisebox{-0.25ex}{\includegraphics[height=1.6ex]{figures/logos/pdf/qwen.pdf}}\,Qwen3.8-27B$^\dagger$ & 38 & 24 & 53 & 5 & 42 & 53 & 90 & 2 & 63 \\
\raisebox{-0.25ex}{\includegraphics[height=1.6ex]{figures/logos/pdf/openai.pdf}}\,GPT-6.1-Sol & 38 & \textbf{58} & 58 & \textbf{32} & 42 & 76 & 92 & 10 & 61 \\
\raisebox{-0.25ex}{\includegraphics[height=1.6ex]{figures/logos/pdf/anthropic.pdf}}\,Claude Opus 5.5 & 38 & 45 & 58 & 21 & 37 & 66 & 90 & 7 & 56 \\
\raisebox{-0.25ex}{\includegraphics[height=1.6ex]{figures/logos/pdf/googlegemini.pdf}}\,Gemini 4 Carbon & 38 & 45 & 68 & 16 & 53 & 55 & 92 & 5 & 55 \\
\raisebox{-0.25ex}{\includegraphics[height=1.6ex]{figures/logos/pdf/xiaohongshu.pdf}}\,dots-3-note$^\dagger$ & 38 & 18 & 37 & 10 & 26 & 55 & 87 & 5 & 52 \\
\raisebox{-0.25ex}{\includegraphics[height=1.6ex]{figures/logos/pdf/nvidia.pdf}}\,Nemotron 3.5 Super VL$^\dagger$ & 38 & 24 & 45 & 0 & 21 & 45 & 87 & 2 & 49 \\
\raisebox{-0.25ex}{\includegraphics[height=1.6ex]{figures/logos/pdf/openai.pdf}}\,GPT-5.6-Sol & 38 & 47 & 58 & 16 & 42 & 66 & 87 & 7 & 46 \\
\raisebox{-0.25ex}{\includegraphics[height=1.6ex]{figures/logos/pdf/googlegemini.pdf}}\,Gemma 4 26B-A4B$^\dagger$ & 38 & 13 & 24 & 0 & 10 & 50 & 87 & 2 & 46 \\
\raisebox{-0.25ex}{\includegraphics[height=1.6ex]{figures/logos/pdf/internlm.pdf}}\,InternVL3.5-38B$^\dagger$ & 38 & 21 & 37 & 5 & 26 & 50 & 82 & 2 & 46 \\
\raisebox{-0.25ex}{\includegraphics[height=1.6ex]{figures/logos/pdf/qwen.pdf}}\,Qwen3.6-35B-A3B$^\dagger$ & 38 & 18 & 37 & 0 & 16 & 53 & 84 & 4 & 46 \\
\raisebox{-0.25ex}{\includegraphics[height=1.6ex]{figures/logos/pdf/anthropic.pdf}}\,Claude Sonnet 4.5 & 38 & 26 & 45 & 5 & 21 & 55 & 84 & 1 & 42 \\
\raisebox{-0.25ex}{\includegraphics[height=1.6ex]{figures/logos/pdf/anthropic.pdf}}\,Claude Haiku 4.5 & 38 & 16 & 29 & 0 & 16 & 47 & 84 & 0 & 39 \\
\raisebox{-0.25ex}{\includegraphics[height=1.6ex]{figures/logos/pdf/openai.pdf}}\,GPT-6-Astra & 38 & 53 & 55 & \textbf{32} & 42 & 74 & 87 & 10 & 38 \\
\raisebox{-0.25ex}{\includegraphics[height=1.6ex]{figures/logos/pdf/qwen.pdf}}\,Qwen3-VL-32B$^\dagger$ & 38 & 8 & 34 & 0 & 16 & 29 & 71 & 0 & 38 \\
\raisebox{-0.25ex}{\includegraphics[height=1.6ex]{figures/logos/pdf/nvidia.pdf}}\,Nemotron 3 Nano Omni$^\dagger$ & 38 & 13 & 53 & 0 & 32 & 45 & 87 & 2 & 36 \\
\raisebox{-0.25ex}{\includegraphics[height=1.6ex]{figures/logos/pdf/googlegemini.pdf}}\,Gemma 4 31B$^\dagger$ & 38 & 24 & 40 & 5 & 21 & 50 & 84 & 2 & 33 \\
\raisebox{-0.25ex}{\includegraphics[height=1.6ex]{figures/logos/pdf/meta.pdf}}\,Muse Glimmer 30B$^\dagger$ & 38 & 16 & 34 & 5 & 16 & 53 & 84 & 5 & 33 \\
\raisebox{-0.25ex}{\includegraphics[height=1.6ex]{figures/logos/pdf/mistralai.pdf}}\,Mistral-Small-3.2-24B$^\dagger$ & 38 & 10 & 34 & 0 & 21 & 34 & 68 & 0 & 26 \\
\raisebox{-0.25ex}{\includegraphics[height=1.6ex]{figures/logos/pdf/qwen.pdf}}\,Qwen3-VL-8B-Thinking$^\dagger$ & 38 & 13 & 24 & 0 & 10 & 37 & 55 & 0 & 17 \\
\raisebox{-0.25ex}{\includegraphics[height=1.6ex]{figures/logos/pdf/kimi.pdf}}\,Kimi-VL-A3B-Thinking$^\dagger$ & 38 & 18 & 21 & 0 & 5 & 45 & 47 & 1 & 8 \\
\raisebox{-0.25ex}{\includegraphics[height=1.6ex]{figures/logos/pdf/qwen.pdf}}\,Qwen3-VL-8B-Instruct$^\dagger$ & 38 & 21 & 21 & 10 & 16 & 47 & 50 & 1 & 6 \\
\raisebox{-0.25ex}{\includegraphics[height=1.6ex]{figures/logos/pdf/qwen.pdf}}\,Qwen3-VL-30B-A3B$^\dagger$ & 38 & 10 & 16 & 0 & 5 & 16 & 29 & 0 & 3 \\
\raisebox{-0.25ex}{\includegraphics[height=1.6ex]{figures/logos/pdf/qwen.pdf}}\,Qwen2.5-VL-7B$^\dagger$ & 38 & 18 & 18 & 0 & 0 & 50 & 45 & 0 & 1 \\
\bottomrule
\end{tabular}

\end{table*}

\FloatBarrier
\section{Full results along the hint ladder}
\label{sec:full}

\Cref{tab:supp-ladder-1k,tab:supp-ladder-m1} give exact accuracy and deciding-person accuracy at every rung and in every mode that each model ran, on \dbench-124 and on \dbench-38.
A dash marks a rung that the model did not run or that is still running; an asterisk on $n$ marks a run in progress, for which only H0 and H3, the rungs run first, are shown.
The last column gives the largest share of unparsed answers in the row, which count as wrong.
The main text summarizes these tables in \cref{tab:main} and \cref{fig:ladder}.

\begin{table*}[!htbp]
  \centering
  \caption{\dbench-124, the subset of \dbench-1K (\NPairsOneForty{} pairs, \NOneForty{} items): accuracy (\%) at each hint rung, by input mode. $n$: items answered in that mode. $^*$Run in progress, H0 and H3 only. Unparsed: the largest share (\%) of unparsed answers across the row's rungs. A dash: not run, or still running. $^\dagger$Open weights. Bold: best in each column (baselines and runs in progress excluded). Data: \dbench-124.}
  \label{tab:supp-ladder-1k}
  \small
  \begin{tabular}{l l r rrrr rrrr r}
\toprule
 & & & \multicolumn{4}{c}{Exact} & \multicolumn{4}{c}{Deciding person} & \\
\cmidrule(lr){4-7}\cmidrule(lr){8-11}
Model & Mode & $n$ & H0 & H1 & H2 & H3 & H0 & H1 & H2 & H3 & Unparsed \\
\midrule
Claude Opus 5.5 & image & 124 & 43 & 44 & 47 & 61 & 59 & 60 & 65 & 91 & 0 \\
Claude Opus 5.5 & text & 124 & 46 & 40 & 47 & 64 & 64 & 60 & 68 & 86 & 0 \\
Claude Sonnet 5.5 & image & 124 & 41 & 40 & \textbf{52} & 60 & 58 & 60 & \textbf{73} & 86 & 0 \\
Claude Sonnet 5.5 & text & 124 & 40 & \textbf{45} & 48 & 58 & 60 & \textbf{66} & 70 & 86 & 0 \\
Claude Haiku 5.5 & image & 124 & 36 & 35 & 39 & 52 & 57 & 59 & 59 & 82 & 0 \\
Claude Haiku 5.5 & text & 124 & 37 & 34 & 40 & 45 & 60 & 53 & 61 & 79 & 0 \\
Claude Opus 4.1 & image & 124 & 42 & -- & -- & 66 & 59 & -- & -- & 88 & 0 \\
Claude Sonnet 4.5 & image & 124 & 18 & -- & -- & 40 & 48 & -- & -- & 80 & 0 \\
Claude Haiku 4.5 & image & 124 & 15 & -- & -- & 25 & 52 & -- & -- & 75 & 0 \\
Claude Fable 5.2 & image & 124 & 54 & -- & -- & 64 & 69 & -- & -- & 85 & 0 \\
GPT-5.6-Sol & image & 124 & 41 & -- & -- & 60 & 57 & -- & -- & 90 & 0 \\
GPT-6-Astra & image & 124 & 59 & -- & -- & 64 & 70 & -- & -- & 83 & 0 \\
GPT-6.1-Sol & image & 124 & \textbf{60} & -- & -- & 62 & \textbf{72} & -- & -- & 82 & 0 \\
Gemini 4 Hydrogen & image & 124 & 47 & -- & -- & \textbf{72} & 62 & -- & -- & 90 & 0 \\
Gemini 4 Carbon & image & 124 & 52 & -- & -- & 64 & 69 & -- & -- & 88 & 0 \\
Gemini 4 Argon & image & 124 & 52 & -- & -- & 69 & 71 & -- & -- & \textbf{92} & 0 \\
Gemini 3.8 Flash & image & 124 & 30 & -- & -- & 61 & 52 & -- & -- & 87 & 0 \\
Gemma 4 31B$^\dagger$ & image & 124 & 22 & -- & -- & 32 & 53 & -- & -- & 69 & 0 \\
Gemma 4 26B-A4B$^\dagger$ & image & 124 & 16 & -- & -- & 33 & 50 & -- & -- & 75 & 0 \\
Nemotron 3.5 Super VL$^\dagger$ & image & 124 & 26 & -- & -- & 46 & 52 & -- & -- & 80 & 0 \\
Nemotron 3 Nano Omni$^\dagger$ & image & 124 & 20 & -- & -- & 30 & 54 & -- & -- & 67 & 0 \\
dots-3-note$^\dagger$ & image & 124 & 29 & -- & -- & 45 & 58 & -- & -- & 79 & 0 \\
Qwen2.5-VL-7B$^\dagger$ & image & 124 & 22 & 18 & 22 & 24 & 52 & 51 & 49 & 50 & 3 \\
Qwen2.5-VL-7B$^\dagger$ & text & 124 & 18 & 15 & 24 & 25 & 48 & 47 & 51 & 50 & 2 \\
Qwen3-VL-8B-Instruct$^\dagger$ & image & 124 & 18 & 11 & 10 & 14 & 53 & 39 & 30 & 44 & 48 \\
Qwen3-VL-8B-Instruct$^\dagger$ & text & 124 & 2 & 6 & 5 & 7 & 11 & 15 & 12 & 17 & 82 \\
Qwen3-VL-8B-Thinking$^\dagger$ & image & 124 & 18 & -- & -- & 18 & 45 & -- & -- & 56 & 23 \\
Qwen3-VL-32B$^\dagger$ & image & 124 & 23 & 23 & 20 & 26 & 60 & 52 & 49 & 73 & 1 \\
Qwen3-VL-32B$^\dagger$ & text & 124 & 26 & 19 & 24 & 27 & 55 & 52 & 52 & 56 & 1 \\
Qwen3-VL-30B-A3B$^\dagger$ & image & 124 & 13 & 6 & 6 & 10 & 26 & 14 & 15 & 20 & 75 \\
Qwen3-VL-30B-A3B$^\dagger$ & text & 124 & 10 & 14 & 4 & 10 & 27 & 27 & 14 & 16 & 73 \\
Qwen3.8-27B$^\dagger$ & image & 124 & 24 & -- & -- & 47 & 47 & -- & -- & 82 & 5 \\
Qwen3.6-35B-A3B$^\dagger$ & image & 124 & 18 & -- & -- & 33 & 44 & -- & -- & 66 & 12 \\
Muse Glimmer 30B$^\dagger$ & image & 124 & 30 & -- & -- & 47 & 55 & -- & -- & 80 & 0 \\
InternVL3.5-38B$^\dagger$ & image & 124 & 12 & -- & -- & 23 & 45 & -- & -- & 69 & 2 \\
Mistral-Small-3.2-24B$^\dagger$ & image & 124 & 17 & 17 & 14 & 23 & 50 & 45 & 48 & 65 & 2 \\
Mistral-Small-3.2-24B$^\dagger$ & text & 124 & 16 & 18 & 22 & 23 & 52 & 52 & 52 & 56 & 2 \\
Kimi-VL-A3B-Thinking$^\dagger$ & image & 124 & 19 & 21 & 20 & 22 & 47 & 45 & 44 & 55 & 6 \\
Kimi-VL-A3B-Thinking$^\dagger$ & text & 124 & 10 & 14 & 15 & 17 & 30 & 33 & 33 & 36 & 33 \\
\bottomrule
\end{tabular}

\end{table*}

\begin{table*}[!htbp]
  \centering
  \caption{\dbench-38 (\NPairsFifty{} pairs, \NFifty{} items): accuracy (\%) at each hint rung, by input mode. $n$: items answered in that mode. Unparsed: the largest share (\%) of unparsed answers across the row's rungs. $^\dagger$Open weights. Bold: best in each column (baselines and runs in progress excluded). Data: \dbench-38.}
  \label{tab:supp-ladder-m1}
  \small
  \begin{tabular}{l l r rrrr rrrr r}
\toprule
 & & & \multicolumn{4}{c}{Exact} & \multicolumn{4}{c}{Deciding person} & \\
\cmidrule(lr){4-7}\cmidrule(lr){8-11}
Model & Mode & $n$ & H0 & H1 & H2 & H3 & H0 & H1 & H2 & H3 & Unparsed \\
\midrule
Claude Opus 5.5 & image & 38 & 45 & \textbf{50} & 42 & 58 & 66 & 68 & 66 & 90 & 0 \\
Claude Opus 5.5 & text & 38 & 40 & 45 & 47 & 63 & 58 & 60 & 68 & 92 & 0 \\
Claude Sonnet 5.5 & image & 38 & 42 & 47 & \textbf{58} & 55 & 58 & 66 & \textbf{82} & 95 & 0 \\
Claude Haiku 5.5 & image & 38 & 34 & 29 & 37 & 53 & 63 & 53 & 60 & 95 & 0 \\
Claude Opus 4.1 & image & 38 & 42 & 40 & 47 & 60 & 66 & 63 & 71 & 95 & 0 \\
Claude Sonnet 4.5 & image & 38 & 26 & 21 & 18 & 45 & 55 & 53 & 50 & 84 & 0 \\
Claude Haiku 4.5 & image & 38 & 16 & 18 & 18 & 29 & 47 & 42 & 50 & 84 & 0 \\
Claude Fable 5.2 & image & 38 & 50 & -- & -- & 55 & \textbf{79} & -- & -- & 87 & 0 \\
GPT-5.6-Sol & image & 38 & 47 & 47 & 55 & 58 & 66 & \textbf{74} & 74 & 87 & 0 \\
GPT-5.6-Sol & text & 38 & 40 & 40 & 37 & 66 & 60 & 60 & 53 & 95 & 0 \\
GPT-6-Astra & image & 38 & 53 & -- & -- & 55 & 74 & -- & -- & 87 & 0 \\
GPT-6.1-Sol & image & 38 & \textbf{58} & -- & -- & 58 & 76 & -- & -- & 92 & 0 \\
Gemini 3.8 Flash & image & 38 & 29 & -- & -- & 58 & 53 & -- & -- & 92 & 0 \\
Gemini 4 Hydrogen & image & 38 & 47 & -- & -- & 63 & 55 & -- & -- & 92 & 0 \\
Gemini 4 Carbon & image & 38 & 45 & -- & -- & 68 & 55 & -- & -- & 92 & 0 \\
Gemini 4 Argon & image & 38 & 53 & -- & -- & \textbf{79} & 71 & -- & -- & \textbf{100} & 0 \\
Gemma 4 31B$^\dagger$ & image & 38 & 24 & -- & -- & 40 & 50 & -- & -- & 84 & 0 \\
Gemma 4 26B-A4B$^\dagger$ & image & 38 & 13 & -- & -- & 24 & 50 & -- & -- & 87 & 0 \\
Nemotron 3.5 Super VL$^\dagger$ & image & 38 & 24 & 26 & 24 & 45 & 45 & 58 & 50 & 87 & 0 \\
Nemotron 3.5 Super VL$^\dagger$ & text & 38 & 18 & 18 & 32 & 37 & 53 & 47 & 60 & 82 & 0 \\
Nemotron 3 Nano Omni$^\dagger$ & image & 38 & 13 & 13 & 32 & 53 & 45 & 40 & 50 & 87 & 0 \\
Nemotron 3 Nano Omni$^\dagger$ & text & 38 & 18 & 24 & 13 & 32 & 42 & 47 & 58 & 76 & 0 \\
dots-3-note$^\dagger$ & image & 38 & 18 & 5 & 10 & 37 & 55 & 16 & 24 & 87 & 63 \\
dots-3-note$^\dagger$ & text & 38 & 10 & 3 & 3 & 24 & 34 & 16 & 5 & 55 & 92 \\
Qwen2.5-VL-7B$^\dagger$ & image & 38 & 18 & 16 & 26 & 18 & 50 & 53 & 53 & 45 & 3 \\
Qwen2.5-VL-7B$^\dagger$ & text & 38 & 24 & 24 & 24 & 29 & 53 & 47 & 55 & 53 & 0 \\
Qwen3-VL-8B-Instruct$^\dagger$ & image & 38 & 21 & -- & -- & 21 & 47 & -- & -- & 50 & 26 \\
Qwen3-VL-8B-Thinking$^\dagger$ & image & 38 & 13 & 16 & 16 & 24 & 37 & 63 & 47 & 55 & 26 \\
Qwen3-VL-8B-Thinking$^\dagger$ & text & 38 & 21 & 21 & 10 & 32 & 42 & 45 & 47 & 74 & 21 \\
Qwen3-VL-32B$^\dagger$ & image & 38 & 8 & -- & -- & 34 & 29 & -- & -- & 71 & 0 \\
Qwen3-VL-30B-A3B$^\dagger$ & image & 38 & 10 & -- & -- & 16 & 16 & -- & -- & 29 & 63 \\
Qwen3.8-27B$^\dagger$ & image & 38 & 24 & -- & -- & 53 & 53 & -- & -- & 90 & 8 \\
Qwen3.6-35B-A3B$^\dagger$ & image & 38 & 18 & -- & -- & 37 & 53 & -- & -- & 84 & 8 \\
Muse Glimmer 30B$^\dagger$ & image & 38 & 16 & -- & -- & 34 & 53 & -- & -- & 84 & 0 \\
InternVL3.5-38B$^\dagger$ & image & 38 & 21 & -- & -- & 37 & 50 & -- & -- & 82 & 0 \\
Mistral-Small-3.2-24B$^\dagger$ & image & 38 & 10 & -- & -- & 34 & 34 & -- & -- & 68 & 3 \\
Kimi-VL-A3B-Thinking$^\dagger$ & image & 38 & 18 & -- & -- & 21 & 45 & -- & -- & 47 & 8 \\
\bottomrule
\end{tabular}

\end{table*}

\FloatBarrier
\section{Identical answers to both twins}
\label{sec:same}

\begin{figure}[!htbp]
  \centering
  \includegraphics[width=0.85\textwidth]{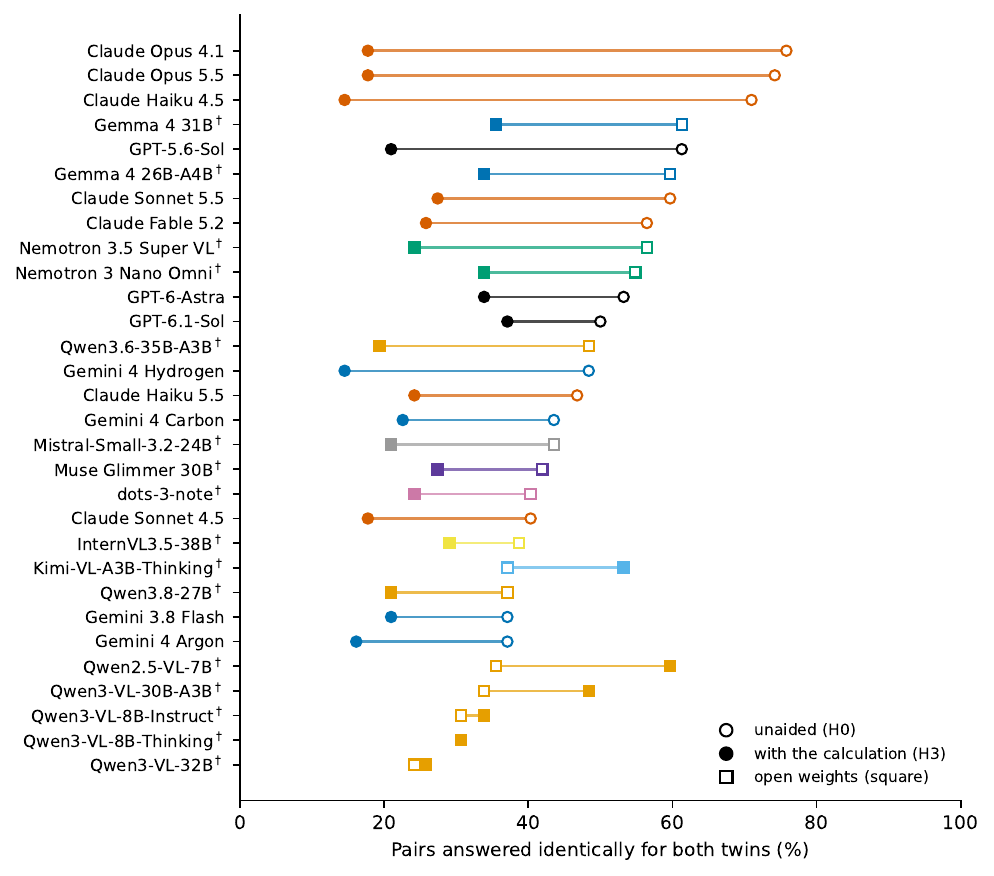}
  \caption{Share of twin pairs answered identically for both twins, unaided (H0, open) and with the deciding calculation (H3, filled), image mode, every model with a complete run. Colour: provider; squares and $^\dagger$: open weights. Data: \dbench-124.}
  \label{fig:supp-same}
\end{figure}

A model that gives the same answer to both twins of a pair is wrong on one of them, since the fate of one person differs between the twins.
\Cref{fig:supp-same} shows, for every model with a complete run on \dbench-124, the share of pairs answered identically.
Unaided, the closed models answer both twins alike in \SameClosedHZeroLo--\SameClosedHZeroHi\% of pairs; with the calculation, in \SameClosedHThreeLo--\SameClosedHThreeHi\%.
Most open-weight models follow the same direction, while \SameUpModels{} answer both twins alike more often with the calculation than without it.

\FloatBarrier
\section{Results by mechanism}
\label{sec:mech}

\Cref{tab:supp-mech} breaks deciding-person accuracy down by the mechanism in which the fate flips, for every model with a complete run on \dbench-1K or \dbench-124, in image mode, unaided (H0) and with the deciding calculation (H3).
For models that answered all items each cell rests on \MechItemsOneKLo{} (the ledge) to \MechItemsOneKHi{} items, and for models that completed \dbench-124 on \MechItemsOneFortyLo{} to \MechItemsOneFortyHi.
\Cref{fig:mechanism} plots every model that answered all items of \dbench-1K.

\begin{table*}[!htbp]
  \centering
  \caption{Deciding-person accuracy (\%) by deciding mechanism, image mode, unaided (H0) and with the deciding calculation (H3). Items: all \NItems{} items of \dbench-1K, or the \NOneForty{} items of \dbench-124. $^\dagger$Open weights. Bold: best in each column; rows with \NOneForty{} items rest on \MechItemsOneFortyLo{} to \MechItemsOneFortyHi{} items per cell and are not comparable with the \NItems-item rows. Data: \dbench-1K and \dbench-124.}
  \label{tab:supp-mech}
  \scriptsize
  \setlength{\tabcolsep}{3pt}
  \begin{tabular}{l l rr rr rr rr rr rr rr}
\toprule
 & & \multicolumn{2}{c}{Ledge} & \multicolumn{2}{c}{Pit} & \multicolumn{2}{c}{Pendulum} & \multicolumn{2}{c}{Recess} & \multicolumn{2}{c}{Seesaw} & \multicolumn{2}{c}{Dominoes} & \multicolumn{2}{c}{Jump} \\
\cmidrule(lr){3-4}\cmidrule(lr){5-6}\cmidrule(lr){7-8}\cmidrule(lr){9-10}\cmidrule(lr){11-12}\cmidrule(lr){13-14}\cmidrule(lr){15-16}
Model & Items & H0 & H3 & H0 & H3 & H0 & H3 & H0 & H3 & H0 & H3 & H0 & H3 & H0 & H3 \\
\midrule
Claude Opus 5.5 & 854 & 82 & 98 & 70 & 93 & 51 & 68 & 57 & 94 & 61 & 99 & 49 & 78 & 70 & 72 \\
Claude Sonnet 5.5 & 854 & 70 & 91 & 62 & 93 & 52 & 61 & 49 & 94 & 55 & \textbf{100} & 54 & 95 & 68 & 71 \\
Claude Haiku 5.5 & 854 & 68 & 68 & 62 & 84 & 57 & 66 & 51 & 94 & 55 & 98 & 55 & 80 & 71 & 71 \\
Claude Opus 4.1 & 854 & 70 & \textbf{100} & 65 & 96 & 51 & 66 & 54 & 94 & 61 & 98 & 50 & 77 & 76 & 71 \\
Claude Sonnet 4.5 & 124 & 43 & 79 & 25 & 60 & 43 & 64 & 60 & 95 & 44 & \textbf{100} & 60 & 95 & 56 & 61 \\
Claude Haiku 4.5 & 124 & 57 & 64 & 50 & 60 & 57 & 50 & 55 & 95 & 44 & \textbf{100} & 50 & \textbf{100} & 56 & 44 \\
Claude Fable 5.2 & 124 & 86 & \textbf{100} & \textbf{80} & 95 & 57 & 79 & 85 & \textbf{100} & 50 & 83 & 50 & 65 & 78 & 72 \\
GPT-5.6-Sol & 854 & 74 & 76 & 72 & 92 & 52 & \textbf{93} & 56 & 96 & 58 & 99 & 52 & 76 & 59 & \textbf{75} \\
GPT-6-Astra & 854 & 85 & 98 & 78 & 93 & 65 & 55 & 91 & 94 & 66 & 98 & 49 & 62 & \textbf{93} & 72 \\
GPT-6.1-Sol & 854 & \textbf{89} & 94 & 74 & 92 & 67 & 60 & \textbf{93} & 95 & 67 & 98 & 52 & 60 & 91 & 74 \\
Gemini 4 Hydrogen & 124 & 79 & \textbf{100} & 55 & \textbf{100} & 86 & 86 & 45 & 95 & 44 & 94 & 55 & 90 & 83 & 67 \\
Gemini 4 Carbon & 124 & 64 & 93 & \textbf{80} & 75 & \textbf{93} & 71 & 60 & \textbf{100} & 50 & \textbf{100} & 60 & \textbf{100} & 83 & 72 \\
Gemini 4 Argon & 124 & 64 & \textbf{100} & 65 & 95 & 64 & 71 & 70 & \textbf{100} & \textbf{78} & \textbf{100} & \textbf{65} & \textbf{100} & 89 & 72 \\
Gemini 3.8 Flash & 124 & 43 & 79 & 55 & 80 & 50 & 79 & 35 & 95 & 61 & \textbf{100} & \textbf{65} & \textbf{100} & 50 & 72 \\
Gemma 4 31B$^\dagger$ & 124 & 50 & 64 & 55 & 55 & 57 & 57 & 50 & \textbf{100} & 56 & 83 & 55 & 70 & 50 & 50 \\
Gemma 4 26B-A4B$^\dagger$ & 854 & 52 & 63 & 55 & 61 & 53 & 74 & 49 & 89 & 48 & 82 & 49 & 97 & 50 & 56 \\
Nemotron 3.5 Super VL$^\dagger$ & 854 & 50 & 54 & 52 & 88 & 52 & 54 & 51 & 88 & 52 & 99 & 48 & 90 & 51 & 62 \\
Nemotron 3 Nano Omni$^\dagger$ & 124 & 43 & 57 & 50 & 45 & 50 & 71 & 65 & 90 & 50 & 83 & 50 & 85 & 67 & 33 \\
dots-3-note$^\dagger$ & 854 & 59 & 57 & 49 & 84 & 49 & 53 & 51 & 98 & 58 & 98 & 42 & 79 & 54 & 70 \\
Qwen2.5-VL-7B$^\dagger$ & 854 & 46 & 57 & 52 & 58 & 40 & 56 & 51 & 45 & 53 & 60 & 47 & 52 & 42 & 52 \\
Qwen3-VL-8B-Instruct$^\dagger$ & 854 & 52 & 35 & 45 & 28 & 49 & 53 & 52 & 46 & 47 & 74 & 36 & 62 & 44 & 6 \\
Qwen3-VL-8B-Thinking$^\dagger$ & 124 & 43 & 57 & 45 & 45 & 36 & 71 & 50 & 65 & 61 & 56 & 30 & 65 & 50 & 33 \\
Qwen3-VL-32B$^\dagger$ & 124 & 50 & 64 & 50 & 45 & 64 & 29 & 55 & 65 & 56 & \textbf{100} & \textbf{65} & 90 & 50 & 50 \\
Qwen3-VL-30B-A3B$^\dagger$ & 854 & 20 & 7 & 29 & 3 & 28 & 3 & 24 & 19 & 28 & 46 & 34 & 18 & 34 & 4 \\
Qwen3.8-27B$^\dagger$ & 854 & 57 & 63 & 55 & 89 & 56 & 57 & 44 & 91 & 52 & 95 & 46 & 84 & 48 & 65 \\
Qwen3.6-35B-A3B$^\dagger$ & 124 & 43 & 43 & 45 & 65 & 43 & 57 & 45 & 80 & 33 & 89 & 50 & 95 & 44 & 22 \\
Muse Glimmer 30B$^\dagger$ & 854 & 50 & 52 & 48 & 70 & 55 & 62 & 54 & 96 & 52 & 98 & 52 & 85 & 56 & 67 \\
InternVL3.5-38B$^\dagger$ & 124 & 36 & 57 & 50 & 55 & 43 & 57 & 50 & 85 & 44 & 94 & 50 & 85 & 39 & 44 \\
Mistral-Small-3.2-24B$^\dagger$ & 124 & 50 & 64 & 55 & 50 & 64 & 64 & 45 & 65 & 44 & 94 & 45 & 85 & 50 & 33 \\
Kimi-VL-A3B-Thinking$^\dagger$ & 854 & 52 & 54 & 48 & 44 & 52 & 68 & 53 & 49 & 45 & 58 & 51 & 68 & 49 & 50 \\
\bottomrule
\end{tabular}

\end{table*}

\FloatBarrier
\section{Mechanisms and their idealized calculations}
\label{sec:rules}

Each mechanism is a part with continuous placement parameters, and a minimal pair varies one of them.
For each mechanism, an idealized calculation predicts the fate of the person it hosts from the speed at which the chain arrives and from quantities that can be read from the grid.
A pair enters the benchmark only if this calculation agrees with the simulator on both twins, and the H3 hint states the calculation.
Geometry is exact for the idealized model: point contact, rolling without slipping, no air drag, gravity $g = 9.81$ squares/s$^2$, people (their collision boxes) 0.6 squares wide and 1.8 tall (0.4 when lying down).

\paragraph{Ledge.}
The floor drops by a height $d$, and a person stands on the lower floor at a distance $s$ from the drop wall.
The ball leaves the edge horizontally at speed $v$, with its centre at height $r$ above the upper floor.
The person dies if the ball's centre comes within $r$ of the person's body at any point of the flight, or if the ball lands short of the person's far side and rolls into them.
Pairs vary $s$ or $d$.

\paragraph{Pit.}
A pit of width $w$ and depth 1.6 holds a crouching person.
The ball leaves the near rim horizontally at speed $v$.
It clears the pit if, when it first touches the far rim's corner, its centre is above the corner, and otherwise it drops in.
Pairs vary $w$.

\paragraph{Pendulum.}
A stone bob hangs on a rigid rod of length $L$, and a person stands at a distance $x_a$ beyond the bob's resting edge.
After the strike the bob moves at speed $v_b$ and rises by $h = v_b^2 / 2g$, so its centre swings sideways by $\sqrt{L^2 - (L - h)^2}$, or by $L$ if it would pass the horizontal.
The person dies if this swing reaches $x_a$.
Pairs vary $x_a$ or $L$.

\paragraph{Recess.}
A person stands in a cage sunk into the floor, with only the head above it, and a ball of radius $r$ with a crescent notch rolls over the head.
The ball starts with the notch facing up and turns once per $2\pi r$ of travel, so when its centre has travelled a path of length $S$ to the point above the head, it has made $n = S / 2\pi r$ turns and the notch has rotated by $360^\circ n$.
The head is spared only if the notch then lies within a window around straight down. The window depends on the ball radius and is calibrated in simulation, from about $\pm 16^\circ$ for $r = 0.6$ to about $\pm 25^\circ$ for $r = 1.2$.
Pairs vary the position of the cage along the floor, and with it $S$.

\paragraph{Seesaw.}
The floor ends at a drop, and below it a seesaw carries a bucket on its raised left end, downward spikes over a person lying under the left arm, and a stone counterweight on its right end.
The ball falls into the bucket.
The person dies if the ball's static torque about the pivot exceeds the restoring torque of the empty seesaw, which is measured in simulation for the counterweight in place.
Pairs vary the counterweight's width or height.

\paragraph{Dominoes.}
A row of wooden dominoes of height $H$ and thickness 0.25 stands with equal gaps, except one wider gap.
A domino falls onto the next only if the gap is shorter than $H$.
In the terminal variant, the last domino, pivoting on its front edge, reaches a person lying at a distance $a$ beyond it if $\sqrt{H^2 - 0.4^2} \geq a$.
Pairs vary the wider gap, the common gap or $a$.

\paragraph{Jump.}
A kicker launches the ball, and a wall of height $h_w$ stands at a distance $g$ beyond the kicker, and a person stands at a distance $s$ behind the wall.
The ball follows a projectile path from the kicker's lip.
If it meets the wall, it rolls over only when the line from the wall's top corner to its centre is steeper than $45^\circ$, a threshold calibrated in simulation, and then rolls on into the person, and otherwise it stops.
If it clears the wall, the person dies when the ball hits them in flight or lands short of their far side and rolls into them, and survives when it overshoots.
Pairs vary $h_w$, $g$ or $s$.

\FloatBarrier
\section{Pairs removed because a death is not visible}
\label{sec:visible}

Each person has a collision box (0.6 by 1.8 squares for a standing person) that reaches slightly beyond the drawn figure near its corners.
We re-ran every item in the unchanged simulator and measured, for every person, how far each moving body enters the collision box and how far it enters the drawn figure (head, torso and limbs as drawn).
In \VisDropOneK{} of the 500 generated pairs of \dbench-1K and \VisDropFifty{} of the 25 pairs of the small benchmark, at least one labelled death comes from a body that enters the box by more than the death depth of 0.02 squares without touching the figure, so a model that reads the picture exactly would be marked wrong.
Most are at ledges (\VisLedgeOneK{} of the \VisDropOneK{} pairs of \dbench-1K), where a falling ball grazes the box beside a person's head.
We removed these pairs from every scored set (\cref{sec:benchmark}).
The removal changes the rankings little.
Comparing DominoScore at H3 with and without these pairs, the Spearman rank correlation is \VisRhoOneKHThree{} over \VisNModelsOneK{} models on \dbench-1K and \VisRhoOneFortyHThree{} over \VisNModelsOneForty{} models on the medium benchmark (\VisRhoOneFortyHZero{} at H0), and scores change by \VisMeanChangeOneKHThree{} and \VisMeanChangeOneFortyHThree{} points on average, consistent with models answering what the picture shows.
On the small benchmark the ledge and the pendulum keep two pairs each, so its scores move more (Spearman \VisRhoFiftyHThree{}, mean change \VisMeanChangeFiftyHThree{}).
\Cref{tab:supp-visible} lists every model on the medium benchmark before (140 items) and after (\dbench-124) the removal.

\begin{table}[!htbp]
  \centering
  \caption{DominoScore on the medium benchmark before the removal of the pairs whose labelled death is not visible (all: 140 items) and after it (visible: \dbench-124), image mode, H0 and H3. Data: \dbench-124 and the \VisDropOneForty{} removed pairs.}
  \label{tab:supp-visible}
  \footnotesize
  \begin{tabular}{l rr rr}
\toprule
 & \multicolumn{2}{c}{H0} & \multicolumn{2}{c}{H3} \\ \cmidrule(lr){2-3} \cmidrule(lr){4-5}
Model & all & visible & all & visible \\
\midrule
Gemini 3.8 Flash & 3.1 & 3.2 & 62.5 & 62.9 \\
Gemini 4 Hydrogen & 26.6 & 30.6 & 61.1 & 65.9 \\
Gemini 4 Argon & 39.9 & 40.0 & 60.4 & 63.2 \\
Claude Opus 5.5 & 8.6 & 8.5 & 59.9 & 63.8 \\
Gemini 4 Carbon & 40.8 & 42.4 & 57.3 & 59.7 \\
Claude Fable 5.2 & 19.2 & 21.0 & 52.5 & 55.9 \\
Claude Opus 4.1 & 8.9 & 9.1 & 48.8 & 55.4 \\
Claude Sonnet 5.5 & 10.5 & 10.1 & 45.4 & 47.4 \\
Gemma 4 26B-A4B & 5.7 & 3.1 & 44.6 & 41.6 \\
Claude Haiku 5.5 & 12.9 & 8.2 & 43.4 & 47.6 \\
Claude Sonnet 4.5 & 2.1 & 4.5 & 33.9 & 39.7 \\
dots-3-note & 16.2 & 20.2 & 28.0 & 32.2 \\
Qwen3.8-27B & 0.8 & 1.0 & 27.6 & 19.0 \\
Nemotron 3.5 Super VL & 2.2 & 5.8 & 25.3 & 27.5 \\
Muse Glimmer 30B & 10.9 & 12.8 & 24.5 & 23.6 \\
Kimi-VL-A3B-Thinking & 1.4 & 1.4 & 21.1 & 11.4 \\
Nemotron 3 Nano Omni & 3.5 & 4.5 & 17.4 & 19.2 \\
Claude Haiku 4.5 & 2.8 & 7.2 & 17.1 & 10.9 \\
Mistral-Small-3.2-24B & 0.4 & 4.6 & 13.7 & 17.0 \\
Gemma 4 31B & 11.3 & 5.4 & 13.4 & 14.8 \\
Qwen3.6-35B-A3B & 0.0 & 0.0 & 12.8 & 14.3 \\
Qwen3-VL-32B & 18.4 & 19.8 & 10.5 & 8.3 \\
Qwen3-VL-8B-Thinking & 0.6 & 0.6 & 8.7 & 13.4 \\
InternVL3.5-38B & 0.0 & 0.0 & 8.2 & 15.2 \\
Qwen3-VL-8B-Instruct & 4.4 & 5.5 & 2.4 & 6.0 \\
Qwen2.5-VL-7B & 3.1 & 3.2 & 1.0 & 1.1 \\
Qwen3-VL-30B-A3B & 0.0 & 0.0 & 0.0 & 0.0 \\
\bottomrule
\end{tabular}

\end{table}

\FloatBarrier
\section{Composition of the benchmarks}
\label{sec:stats}

\Cref{tab:supp-stats} describes \dbench-1K, the full benchmark, and \dbench-38, the small benchmark (\dbench-124, the medium benchmark, is a subset of \dbench-1K; \cref{sec:setup} explains which models ran on which benchmark).
\dbench-38 was generated separately with the same checks.
It has no jump mechanism, and its pairs were found by a search over random seeds, while \dbench-1K's pairs were placed by the guided search of \cref{sec:benchmark}.
The pairs of \dbench-1K were selected from the larger pool of solvable candidates so that each mechanism has 71 or 72 pairs; after the removal of the pairs with a death that the picture does not show (\cref{sec:visible}), the ledge keeps \PerMechOneKLo{} and the other mechanisms \PerMechOneKSecondLo{} to \PerMechOneKHi.

\begin{table*}[!htbp]
  \centering
  \caption{Composition of \dbench-1K and \dbench-38. Counts are of items unless the row says pairs. A pair's twin separation is the number of grid steps between the two values of its varied parameter (0.25 squares for lengths, 0.1 for ball radii). Data: \dbench-1K and \dbench-38.}
  \label{tab:supp-stats}
  \small
  \begin{tabular}{p{0.33\textwidth} p{0.37\textwidth} p{0.22\textwidth}}
\toprule
 & \dbench-1K & \dbench-38 \\
\midrule
Pairs / items & 427 / 854 & 19 / 38 \\
Pairs per deciding mechanism & Ledge 27, Pit 65, Pendulum 62, Recess 70, Seesaw 65, Dominoes 68, Jump 70 & Ledge 2, Pit 2, Pendulum 2, Recess 5, Seesaw 5, Dominoes 3, Jump 0 \\
Mechanisms reached in the simulation (items) & 3: 489, 4: 238, 5: 127 & 3: 21, 4: 13, 5: 4 \\
People per item, incl. the pusher (items) & 2: 64, 3: 332, 4: 308, 5: 132, 6: 18 & 2: 2, 3: 10, 4: 16, 5: 10 \\
Deaths per item (items) & 0: 216, 1: 380, 2: 211, 3: 47 & 0: 10, 1: 16, 2: 8, 3: 3, 4: 1 \\
Twin separation in grid steps (pairs) & 1: 89, 2: 208, 3: 130 & 1: 10, 2: 7, 3: 1, 4: 1 \\
Items where ``everyone but the pusher dies'' is wrong & 731 of 854 & -- \\
Solvable candidate pairs before selection & 994 & 36 \\
\bottomrule
\end{tabular}

\end{table*}

\FloatBarrier
\section{Example pairs}
\label{sec:examples}

\begin{figure*}[!htbp]
  \centering
  \includegraphics[width=\textwidth]{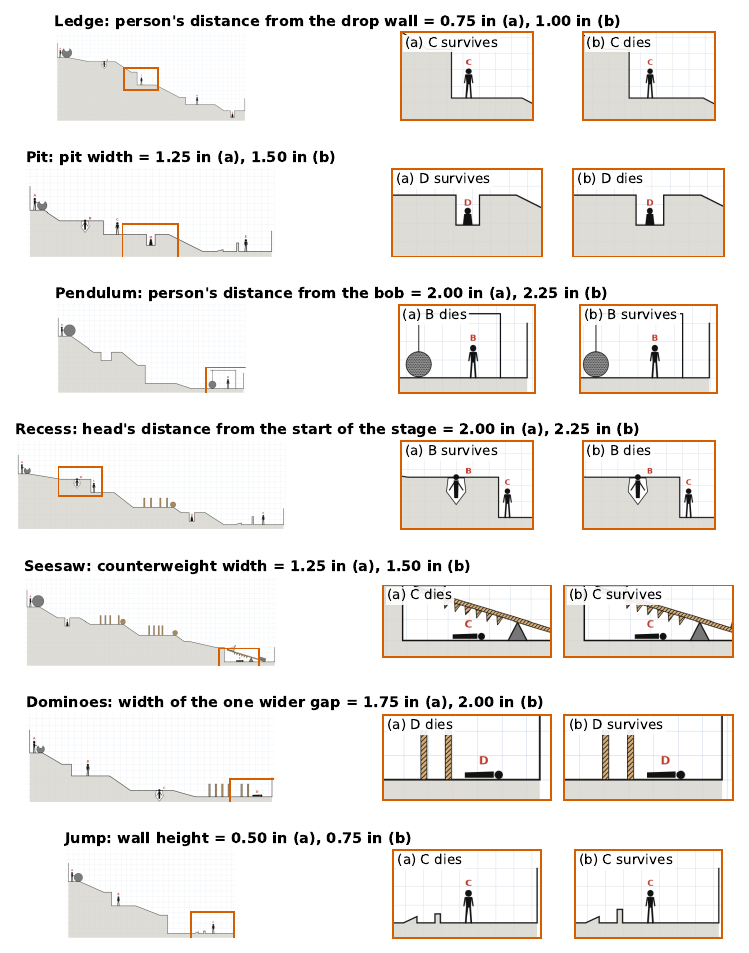}
  \caption{One minimal pair per deciding mechanism. Left: the still of item (a), with the person whose fate flips boxed. Right: that person in item (a) and in item (b), with the simulated fate. The row title names the parameter that differs and its value in each item (in grid squares). Data: \dbench-1K.}
  \label{fig:supp-examples}
\end{figure*}

\Cref{fig:supp-examples} shows one pair for each mechanism: the lowest-numbered pair among those with the smallest twin separation.

\FloatBarrier
\section{Prompts, answer parsing and DominoScore weights}
\label{sec:prompts}

\paragraph{Prompts.}
Every call has a system prompt, which states the rules of the world, and a user prompt, which holds the scene, the question and the hints up to the requested rung.
\Cref{fig:supp-prompts} shows the exact prompts for item 0004-a (the pair of \cref{fig:teaser}): the system prompt, the user prompt in image mode at H3, and the user prompt in text mode at H0.

\begin{figure*}[!htbp]
  \centering
  \begin{minipage}[t]{0.325\textwidth}
    \textbf{\small System prompt.}
\begin{lstlisting}
You are solving a physics puzzle about a chain reaction.

The scene is a 2D side view. Distances are in grid units (1 grid square = 1 unit). Gravity points down at 9.81 units/s^2.
Rules of this world:
- 2D rigid bodies; mass = density x area (stone 2.5, wood 0.6 per square of area)
- balls roll without slipping unless they are skidding after a landing
- gravity 9.81 squares/s^2, pointing down; the push sets the ball rolling to the right at 1.5 squares/s
- a person dies if any moving solid object (ball, domino, plank, spikes, hanging ball) hits them
- people do not move and do not stop objects

Work through the chain of events step by step and check any quantity that could change the outcome. List the assumptions you rely on.
Finish with your final answer as a JSON object in a ```json code block, with exactly these keys:
  "answer": an object mapping every person's letter to "dies" or "survives"
  "p_dies": an object mapping every person's letter to your probability (0 to 1) that they die
  "deciding_factors": one or two sentences naming what decided the outcome
The JSON block must be the last thing you write.
\end{lstlisting}
  \end{minipage}\hfill
  \begin{minipage}[t]{0.325\textwidth}
    \textbf{\small User prompt, image mode, H3.}
\begin{lstlisting}
The attached picture shows the scene, drawn on a grid (1 square = 1 unit), with a legend for materials.

People in the scene: A, B, C.

Question: Who dies if A pushes the ball?

Hint: Before answering, consider the quantitative details in the picture that could change the outcome.

Hint: Consider whether the ball landing in the bucket can outweigh the counterweight above C.

Hint: Compare turning effects (mass x horizontal distance from the pivot). The ball (mass about 0.47: wood, area pi r^2 with r = 0.5) comes to rest in the bucket about 4.35 squares from the pivot. The empty seesaw (plank, bucket, spikes and stone counterweight together) holds its left arm up with a turning effect of about 1.85. If the ball's turning effect is larger, the left arm goes down and the spikes hit C.
\end{lstlisting}
  \end{minipage}\hfill
  \begin{minipage}[t]{0.325\textwidth}
    \textbf{\small User prompt, text mode, H0.}
\begin{lstlisting}
Scene description: A stands on a platform and pushes a solid stone ball, 2.4 squares across, to the right. The floor slopes down 2.25 squares over 3.75 squares. A pit 1.00 squares wide and 1.6 squares deep; B crouches at its bottom. The floor continues at the same height on the far side. The floor slopes down 1.25 squares over 2.25 squares. A row of 4 wooden dominoes, each 1.75 squares tall and 0.25 thick, standing 0.75 squares apart, except a 1.50-square gap between domino 3 and domino 4. 0.25 squares after the last domino a small wooden ball (1.0 across) rests at the top of a slope that descends 2.25 squares. A row of 4 wooden dominoes, each 2.00 squares tall and 0.25 thick, standing 0.75 squares apart, except a 0.50-square gap between domino 3 and domino 4. 2.00 squares after the last domino a small wooden ball (1.0 across) rests at the top of a slope that descends 1.25 squares. The floor slopes down 1.25 squares over 5.50 squares. The floor ends at a 3.00-square drop. Below it is a wooden seesaw: the left arm is 5.25 squares long and ends in a bucket just below the drop; the right arm is 3.00 squares long and carries a stone counterweight block 1.25 squares wide and 0.50 tall, and rests on the ground. Spikes point down from the left arm over C, who lies on the ground.

Exact geometry (units, x to the right, y up):
{"terrain_outline": [[0.0, 14.75], [0.0, 11.25], [2.6, 11.25], [6.35, 9.0], [7.35, 9.0], [7.85, 9.0], [7.85, 7.4], [8.85, 7.4], [8.85, 9.0], [10.85, 9.0], [13.1, 7.75], [14.1, 7.75], [19.93, 7.75], [23.143, 5.5], [24.143, 5.5], [30.723, 5.5], [32.509, 4.25], [33.509, 4.25], [39.009, 3.0], [40.009, 3.0], [40.809, 3.0], [40.809, 0.0], [50.729, 0.0], [51.229, 0.0], [51.229, 3.0]], "other_static_lines": [[[45.909, 0.0], [46.359, 0.873], [46.809, 0.0]]], "objects": [{"name": "ball0", "kind": "ball", "material": "stone", "position": [2.3, 12.452], "angle_deg": 0.0, "mass": 11.31, "r": 1.2, "notched": false}, {"name": "domino4_0", "kind": "domino", "mat ... [geometry truncated here for the page; the prompt carries it in full]

People in the scene: A, B, C.

Question: Who dies if A pushes the ball?
\end{lstlisting}
  \end{minipage}
  \caption{Exact prompts for item 0004-a (the pair of \cref{fig:teaser}). In image mode the picture is attached to the user prompt; in text mode the exact geometry is a JSON object of several thousand characters, shortened here.}
  \label{fig:supp-prompts}
\end{figure*}

\paragraph{Answer parsing.}
The parser takes the last fenced JSON block in the response, or else the last balanced pair of braces, and reads its \texttt{answer} object.
Person names are normalized to their capital letter, and verdicts beginning with \emph{die}, \emph{dead} or \emph{kill} map to dies, and those beginning with \emph{surv}, \emph{live}, \emph{alive} or \emph{safe} to survives.
An answer is parsed only if it gives a valid verdict for every person in the scene. Any other response, including one cut off by the token budget, counts as wrong.

\paragraph{DominoScore weights.}
\Cref{tab:supp-weights} lists the frozen weights of \cref{eq:dominoscore}.
A mechanism's difficulty is the error rate on the deciding person, at H3, of the four Claude models that answered every item.
The weights are the difficulties normalized to sum to one, with every weight floored at half the uniform weight and renormalized, so the seesaw, the easiest mechanism with the calculation, keeps a weight of \WeightMin. We computed the weights after removing the pairs of \cref{sec:visible}; computed on all 1,000 generated items instead, they differ most for the \WeightMaxChangeMech{} (\WeightMaxChangeOld{} instead of \WeightMaxChangeNew).

\begin{table}[!htbp]
  \centering
  \caption{Frozen DominoScore weights. Difficulty: error rate on the deciding person at H3 of the reference panel. Weight: normalized difficulty, floored at half the uniform weight and then renormalized. Data: \dbench-1K.}
  \label{tab:supp-weights}
  \small
  \begin{tabular}{l rr}
\toprule
Mechanism & Difficulty $d_m$ & Weight $w_m$ \\
\midrule
Pendulum & 0.345 & 0.300 \\
Jump & 0.287 & 0.250 \\
Dominoes & 0.175 & 0.152 \\
Ledge & 0.106 & 0.092 \\
Pit & 0.085 & 0.073 \\
Recess & 0.059 & 0.066 \\
Seesaw & 0.013 & 0.066 \\
\bottomrule
\end{tabular}

\end{table}

\paragraph{DominoScore robustness.}
Because the skill of each mechanism is clipped at zero, a model that guesses scores above zero on a finite sample.
We simulated 20,000 guessers that draw every person's fate with a fair coin, independently for each item, on the actual items of each benchmark (\cref{tab:supp-ds-robustness}).
A guesser scores \RandDSOneKMean{} on average on \dbench-1K (95th percentile \RandDSOneKPct), \RandDSOneFortyMean{} on \dbench-124 (\RandDSOneFortyPct) and \RandDSFiftyMean{} on \dbench-38 (\RandDSFiftyPct), so on the smaller benchmarks low scores cannot be told apart from guessing.
The thresholds of \cref{tab:main} (10 on \dbench-1K and 20 on \dbench-124) lie at or above these percentiles.
The frozen weights come from the four Claude models that answered every item, so we also scored every complete run with equal weights for all mechanisms.
The two rankings at H3 agree closely: Spearman $\rho = \UniformRhoOneK$ over \UniformNOneK{} models on \dbench-1K, \UniformRhoOneForty{} over \UniformNOneForty{} on \dbench-124 and \UniformRhoFifty{} over \UniformNFifty{} on \dbench-38.
To measure sampling noise, we ran Claude Opus 5.5 a second time on \dbench-124 (image mode, H0 and H3, same prompts and settings).
The second run gives the identical answer on \RepSameHZero\% of items at H0 and \RepSameHThree\% at H3.
Its deciding-person accuracy moves from \RepDecHZeroA\% to \RepDecHZeroB\% at H0 and from \RepDecHThreeA\% to \RepDecHThreeB\% at H3, and its DominoScore from \RepDSHThreeA{} to \RepDSHThreeB{} at H3, inside the 95\% interval of the first run, and from \RepDSHZeroA{} to \RepDSHZeroB{} at H0, just above it.
On \dbench-124, differences of up to about \RepDSDiffHThree{} points of DominoScore at H3 between two models are therefore within the noise of a single run.

\begin{table}[!htbp]
  \centering
  \caption{DominoScore of a random guesser (mean and 95th percentile over 20,000 simulations) and the Spearman rank correlation at H3 between DominoScore with the frozen weights and with uniform weights, over every complete run. Data: all three benchmarks.}
  \label{tab:supp-ds-robustness}
  \footnotesize
  \setlength{\tabcolsep}{4pt}
  \begin{tabular}{l rr rr}
\toprule
 & \multicolumn{2}{c}{Random guesser} & \multicolumn{2}{c}{Uniform weights} \\
\cmidrule(lr){2-3}\cmidrule(lr){4-5}
Benchmark & mean & 95th pct. & models & Spearman $\rho$ \\
\midrule
\dbench-1K & 2.6 & 6.2 & 16 & 0.95 \\
\dbench-124 & 5.6 & 14.4 & 30 & 0.96 \\
\dbench-38 & 8.0 & 25.9 & 30 & 0.97 \\
\bottomrule
\end{tabular}

\end{table}

\end{document}